\documentclass{article}

\usepackage[preprint]{corl_2026} 

\usepackage{graphicx}
\usepackage[utf8]{inputenc}
\usepackage{booktabs}
\usepackage{tabularx} 
\usepackage{xcolor}   
\usepackage[table]{xcolor}
\usepackage{pifont}
\usepackage{wrapfig}
\usepackage{titletoc}

\usepackage{tcolorbox}
\tcbuselibrary{breakable}

\usepackage[dvipsnames]{xcolor}

\definecolor{lightblue}{RGB}{100, 149, 237}

\usepackage{colortbl}
\usepackage{multirow}
\usepackage[para]{footmisc}
\usepackage{lipsum}
\usepackage{booktabs}

\usepackage{booktabs}
\usepackage{graphicx}
\usepackage[table]{xcolor}
\usepackage{siunitx}
\usepackage{tcolorbox}

\definecolor{lightgreen}{RGB}{210,250,210}

\usepackage{xspace}
\usepackage{hyperref}

\newcommand{\ms}[2]{$#1{\scriptstyle\,\pm\,#2}$}

\makeatletter
\DeclareRobustCommand\onedot{\futurelet\@let@token\@onedot}
\def\@onedot{\ifx\@let@token.\else.\null\fi\xspace}

\makeatother

\usepackage{amsmath}
\usepackage{amssymb}
\usepackage{dsfont}
\usepackage{etoolbox}
\usepackage{color}
\usepackage{ulem}

\newif\ifshowedits

\newcommand{\addeditor}[3]{%
  \definecolor{#1color}{rgb}{#3}
  \expandafter\newcommand\csname #1\endcsname[1]{%
  \ifshowedits
    {\color{#1color} ##1}%
  \else
    {##1}%
  \fi
  }%
  \expandafter\newcommand\csname #1rmk\endcsname[1]{%
  \ifshowedits
    {\color{#1color} {\bf [#2: ##1]}}
  \fi
  }%
  \expandafter\newcommand\csname #1rpl\endcsname[2]{%
  \ifshowedits
    {\color{#1color} ##1 \sout{##2}}
  \else
    {##1}
  \fi
  }%
}

\newcommand{\createtextvar}[1]{
  \expandafter\newcommand\csname #1\endcsname{%
  {\text{#1}}
}%
}
\newcommand{\mycomment}[1]{}

\newcommand{\vcomment}[1]{}

\title{Robot-WATCHDOG : failure detection through object-centric graph representation}

\author{
Quentin Rolland$^{1,2}$ \quad
Fabrice Mayran de Chamisso$^{1}$ \quad
Jean-Baptiste Mouret$^{2,3}$\\[2mm]
$^{1}$ Université Paris-Saclay, CEA, List, F-91120, Palaiseau, France\\
$^{2}$ Inria, CNRS, Université de Lorraine, LORIA, F-54000 Nancy, France\\
$^{3}$ Bleu Robotics, Paris, France
}

\begin{document}
\maketitle

\addeditor{jeanBaptiste}{JB}{1, 0.4, 0.0}
\addeditor{fabrice}{FAB}{0.6, 0, 1}
\addeditor{quentin}{Q}{0, 0.5, 0.5}

\showeditsfalse

\begin{figure}[h]
    \centering
    \includegraphics[width=13.5cm]{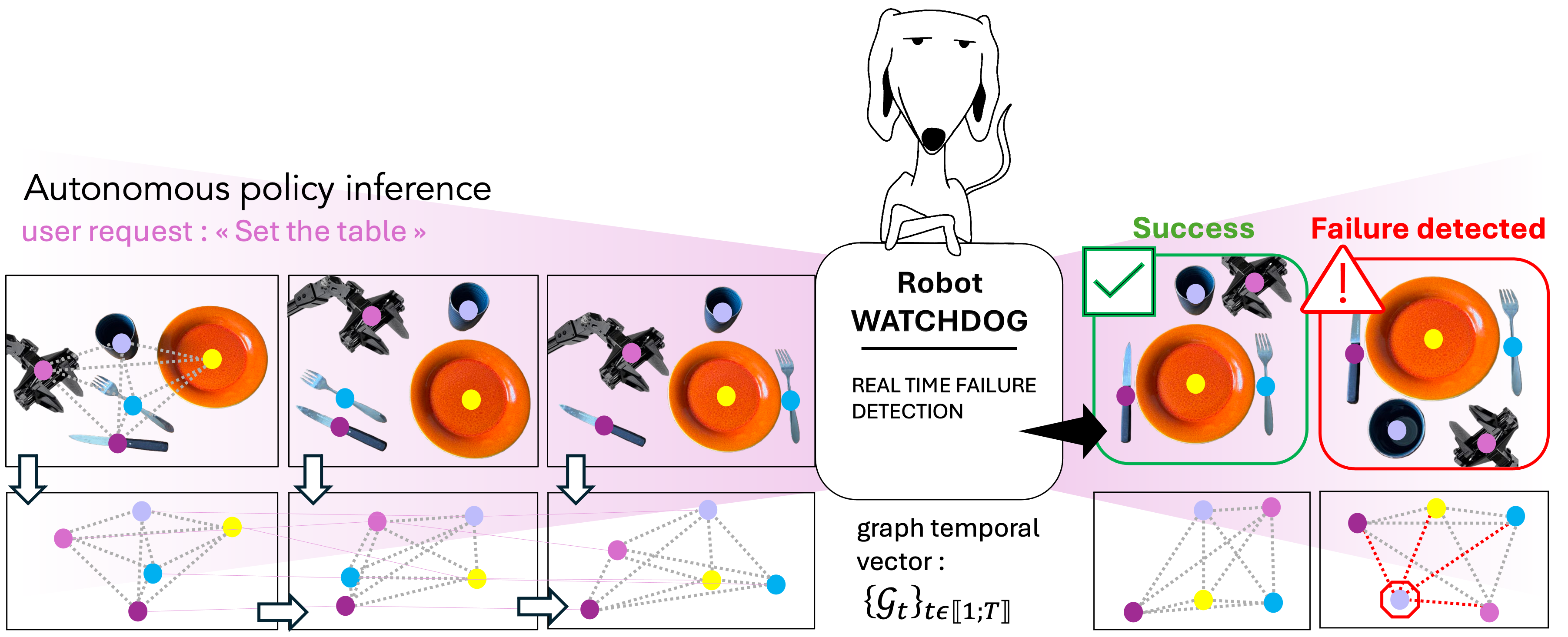}
    \caption{\textbf{Detecting failures through object-centric relational dynamics.}
        Robot-WATCHDOG represents manipulation scenes as spatio-temporal graphs enabling robust detection of relational and temporal errors beyond raw visual or trajectory-based methods.
    }
    \label{fig:intro_fig}
\end{figure}

\begin{abstract}

Reliable real-time failure detection is critical for deploying learned robotic policies, yet defining failures explicitly is infeasible in open-world settings. This motivates failure detection from nominal demonstrations alone. Existing methods, however, struggle to balance robustness to benign variations with the ability to capture semantic and interaction-level errors. We introduce \textbf{Robot-WATCHDOG}, a self-supervised framework that models manipulation scenes as object-centric spatio-temporal graphs. Building on this representation, we investigate two complementary failure scores: \textbf{Trajectory Projection and Correlation (TPC)}, a non-parametric measure of geometric deviation and relational consistency, and \textbf{GnnT}, a Graph Attention Network with a Transformer that predicts future object states.  We evaluate both on a public benchmark and on a new dataset of semantically complex manipulation tasks, across six tasks and ten random seeds. Both scores outperform image-based failure detectors, particularly on semantic and interaction-level failures. TPC provides the strongest performance in AUPR and MCC on both dataset averages, with low variance across seeds. Qualitative results are made available at \url{https://cea-list.github.io/Robot-WATCHDOG/}.

\end{abstract}



	




\section{INTRODUCTION}

Learned robotic policies provide no guarantees beyond the distribution induced by training demonstrations~\cite{agia2026deploymenttimereliabilitylearnedrobot}. Detecting deviations from expected behavior is therefore essential for accountable deployment, and increasingly emphasized in regulation for high-risk AI systems such as the EU AI Act~\cite{HerreraPoyatos2025AFF, valdenegro2024}. As imitation learning and vision-language-conditioned policies increasingly enable long-horizon, closed-loop manipulation~\cite{chi2024diffusion, ACT, rt22023arxiv, kim24openvla, 2026pi07, gr00tn1_2025}, policies are executed autonomously in real time and without step-by-step human intervention, motivating dedicated monitoring systems that operate online, require no failure examples, and can identify anomalous behavior before errors propagate.

Explicitly enumerating all possible failures, however, is infeasible in open-world robotic settings. The standard framing therefore casts failure detection as a one-class problem~\cite{xu2025FailDetect, rolland2026failure, fiper}: a monitor trained exclusively on nominal demonstrations must identify, at runtime, deviations from expected behavior. Existing approaches divide broadly into two families. Visual failure detectors~\cite{Thoduka_2021, gu2025safe, zhou2026rcnfrobot, rolland2026failure} flag deviations in raw image space or learned visual representations. They are sensitive to benign background changes, lighting shifts, and visual clutter, producing frequent false positives that erode operational efficiency. Kinematic and action-based monitors~\cite{Lello2013BayesianTM, grambow2026anomalydetectiongenericfailure, 681254} instead operate on trajectory or joint-space signals; they are robust to visual noise but structurally blind to the environment. They only verify whether the movement itself is familiar, failing to detect if the robot is interacting with the wrong object or operating in an environment that mismatches training.

This division points to a broader limitation. A substantial class of failures in robotic manipulation does not arise from abnormal motion or corrupted observations, but from incorrect interactions between otherwise familiar objects: grasping the wrong item, violating task-order constraints, or placing an object in an incorrect spatial relation to others~\cite{SurveyADfam, ADsurvey2}. These failures emerge naturally in manipulation tasks whose success depends on object identity, relative placement, and sequential consistency, yet they remain difficult to capture when monitoring is restricted to either image appearance or motion trajectories alone.

The central idea of this work is to recast robotic failure detection as monitoring the evolution of object interactions over time. Rather than treating visual and kinematic signals as separate modalities, we represent the scene as a spatio-temporal graph of interacting objects, where nodes encode object states and edges encode their pairwise spatial relationships throughout execution. This object-centric formulation generalizes prior monitoring approaches: visual detectors can be interpreted as operating on node appearance, while kinematic monitors correspond to tracking motion along trajectories. By jointly modeling objects, their relations, and their temporal evolution in a unified representation, the proposed framework captures both the signals targeted by prior work and the relational structure required to detect task-relevant interaction failures.

Failure detection then becomes a well-posed problem: deviations from reference relational dynamics constitute anomalies. We instantiate this principle in Robot-WATCHDOG, a self-supervised framework that abstracts a manipulation episode into an object-centric spatio-temporal graph and scores each timestep against expert demonstrations. Within this representation we study two scoring functions. TPC (Trajectory Projection and Correlation) is non-parametric: it measures orthogonal projection distances to expert trajectories together with deviations in the pairwise correlation structure between objects. GnnT couples a Graph Attention Network with a Transformer encoder to predict future object dynamics from a sliding observation window, and flags prediction errors. Both are converted into binary decisions by conformal prediction calibrated on held-out expert demonstrations, which sets the decision threshold from a user-specified level without tuning the threshold on failure examples.

We evaluate Robot-WATCHDOG on four tasks of the public BotFails benchmark \cite{rolland2026failure} and on a newly introduced dataset featuring two relationally challenging manipulation tasks: structured table setting and waste sorting on a moving conveyor belt. All results are averaged over ten seeds, each redrawing the expert partition and the network initialisation. Both object-centric scores outperform image-based detectors, and TPC is consistently the stronger of the two: averaged per dataset it reaches AUPR 0.81 and maximum MCC 0.75 (BotFails) and 0.81 / 0.72 (Watchdog), against 0.62 / 0.49 and 0.65 / 0.63 for GnnT, while the best baseline reaches only 0.44 / 0.32 and 0.35 / 0.34. After conformal calibration, TPC attains MCC 0.68 and 0.64 where the strongest baseline reaches 0.22 and 0.33. The learned score only leads in the continuously dynamic conveyor scene. Methods that perform well on simple kinematic deviations struggle with relational failures involving object identity and spatial arrangement, the very setting Robot-WATCHDOG addresses.

Our main contributions are: (i) identifying relational anomalies as a key yet underexplored failure mode in robotic manipulation, arising from dependencies on object identity, spatial arrangement, and temporal consistency; (ii) introducing two scoring functions: TPC (Trajectory Projection and Correlation) a non-parametric, geometric method and GnnT, a spatio-temporal graph formulation ; and (iii) releasing a benchmark of relationally challenging manipulation tasks with annotated relational failures.

\section{RELATED WORK}


\subsection{Graph oriented object representation}

A growing line of work in robotics and visuomotor learning has explored object-centric graph representations, where nodes encode task-relevant entities and edges capture spatial, label-conditioned, or action-conditioned relations~\cite{orion, nri, interaction_networks}. Such representations improve invariance to irrelevant visual variations and provide a structured way to model interactions between objects. Building on this paradigm, several approaches construct dynamic or action-conditioned scene graphs to model evolving environments, improving robustness and generalization in manipulation tasks~\cite{roboexp, graphcotvla}. In parallel, graph neural networks have been leveraged for relational policy learning, planning, and dynamics prediction, demonstrating improved scalability in multi-object settings~\cite{gnn_relational_classifiers, objectzero}. More recently, Vosylius et al.~\cite{instantpolicy} show that graph-structured representations can support in-context policy adaptation by modeling demonstrations over relational structures. Recent progress in vision-language models and object tracking has also made these representations increasingly practical in robotics, enabling automatic extraction of task-relevant objects and trajectories directly from RGB observations. To our knowledge, while object-centric graph representations have shown strong promise for policy learning and planning, they have received limited attention for online failure detection, motivating their investigation in this work.

\subsection{Failure detection in autonomous robotics}

Existing real-time monitors struggle to balance environmental awareness, robustness to benign variations, and sensitivity to task-relevant object interactions. Furthermore, many traditional failure classifiers~\cite{gokmen2023askinghelpfailureprediction, liu2023modelbasedruntimemonitoringinteractive, liu2024multitaskinteractiverobotfleet} require collecting explicit failure data \textit{a priori}, a time-consuming process that scales poorly to open-world robotic deployment. Consequently, recent efforts focus on one-class methods trained exclusively on nominal demonstrations. However, vision-based detectors like FIDeL~\cite{rolland2026failure} remain highly sensitive to visual clutter. Sentinel~\cite{agia2024} attempts to bridge the modality gap by combining STAC : a statistical measure of kinematic action consistency, with a Vision-Language Model (VLM) for visual verification. Yet, STAC remains blind to environment-only failures, and the VLM fallback introduces severe inference latency unsuited for reactive continuous control. Similarly, FAIL-Detect~\cite{xu2025FailDetect} merges modalities through latent concatenation, while FIPER~\cite{fiper} relies on brittle logical conjunctions.

Crucially, these approaches often struggle to capture failures involving object identity, spatial arrangement, or task progression, such as grasping the wrong object, executing sub-actions out of order, or misplacing items. Such errors often produce nominal kinematic profiles and only subtle pixel-level changes, making them difficult to detect with action-consistency metrics or standard visual anomaly detectors.

Our framework, Robot-WATCHDOG, addresses these limitations through an object-centric representation of scene dynamics. By abstracting tracked entities into a spatio-temporal graph, the framework reduces sensitivity to irrelevant visual variation while preserving explicit environmental context unavailable to kinematic-only monitors. Representing the scene in a relational geometric space also enables anomaly detection based on changes in object interactions over time, either directly, by comparing object trajectories and their pairwise correlations to expert demonstrations (TPC), or through a predictive GNN-Transformer that jointly models object identities, spatial configurations, and temporal dependencies (GnnT). Both enable real-time detection of relational failure modes that remain challenging for existing multimodal baselines, without requiring failure demonstrations during training.
\section{Method}
\label{sec:method}

\begin{figure}[h]
    \centering
    \includegraphics[width=\textwidth]{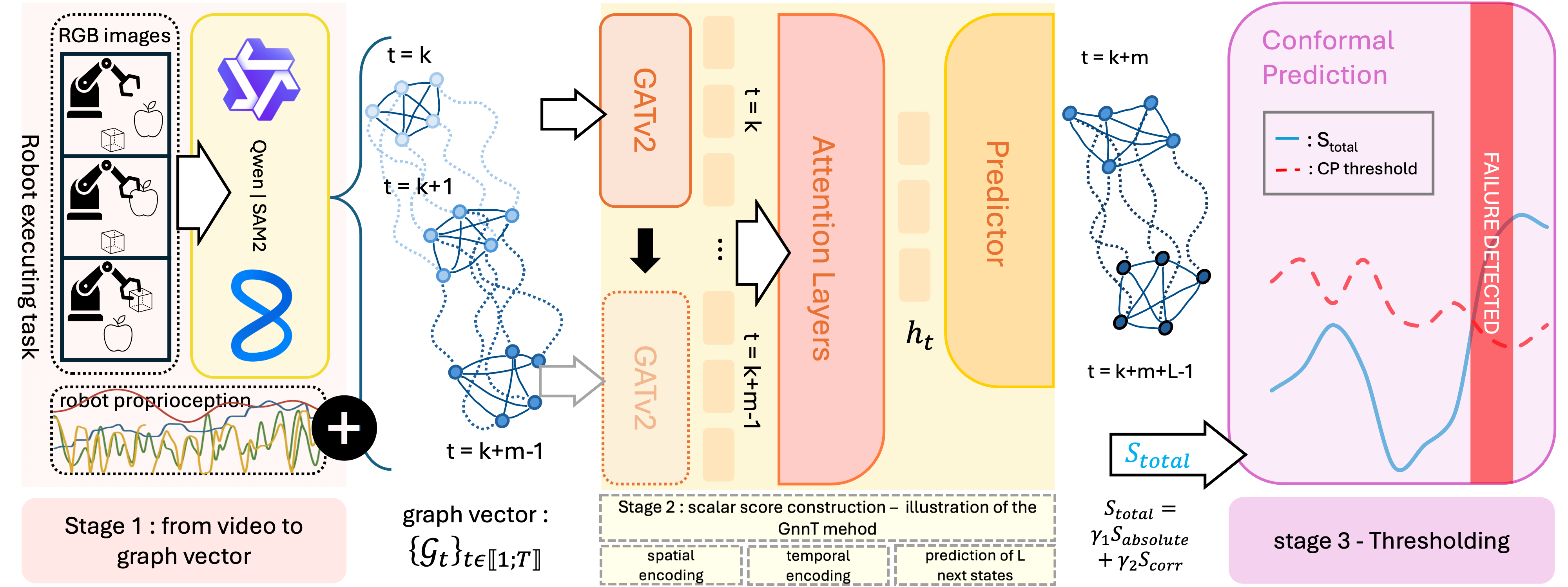}
    \caption{
        Overview of the Robot-WATCHDOG pipeline.
        \textbf{Stage 1:} object-centric graph extraction from RGB observations and robot proprioception. 
        \textbf{Stage 2:} failure detection (illustrated with \textit{GnnT}), which takes the graph as input and outputs a scalar failure score $S_{\text{total}} \in \mathbb{R}$ at each timestep. 
        \textbf{Stage 3:} conformal prediction, where $S_{\text{total}}$ is compared to a threshold calibrated offline on demonstration data, yielding a binary decision (failure / no failure).
    }
    \label{fig:global_fig}
\end{figure}

As illustrated in Figure 2, we propose a self-supervised failure detection framework for multi-object robotic manipulation. The framework rests on a single representation: each episode is abstracted into a sequence of object-centric spatio-temporal graphs built from RGB observations and
robot proprioception. Within this representation we define two scoring functions, both trained or referenced on nominal demonstrations only, and both turned into binary decisions by the same conformal calibration. TPC is non-parametric and scores an episode by its geometric and relational
deviation from expert trajectories. GnnT learns the nominal dynamics and scores an episode by its future-prediction error. At every timestep each scorer produces one value per tracked object; the scalar failure score $S_{\text{total}}$ is their sum over objects.

\subsection{Problem Formulation and Graph Extraction}

We consider robotic manipulation episodes involving $C$ distinct entities (task-relevant objects together with the robot end-effector) evolving in a fixed workspace. Using a vision-language initialization model and a tracking pipeline~\cite{bai2025qwen3vltechnicalreport, ravi2024sam2}, each episode is converted from RGB observations into a sequence of 2D trajectories over tracked labeled objects. This yields a temporal graph sequence $\{\mathcal{G}_t\}_{t=1}^{T}$, which is the common input of both scoring functions defined below.

At each time step $t$, the scene is abstracted as $\mathcal{G}_t=(V,E_t,X_t)$. The vertex features $X_t \in \mathbb{R}^{C\times d}$ of vertices V encode absolute positions $\mathbf{p}_{i,t}$ and instantaneous velocities: $\mathbf{x}_{i,t} = [\mathbf{p}_{i,t}, \mathbf{p}_{i,t}-\mathbf{p}_{i,t-1}]$. Edges $E_t$ encode the explicit pairwise spatial relations. For any pair $(i,j)$, the relative displacement vector $\mathbf{o}_{ij,t} = \mathbf{p}_{j,t} - \mathbf{p}_{i,t}$ yields a Euclidean distance $d_{ij,t}$ and an orientation $\theta_{ij,t}$. The edge feature is defined as : $ \mathbf{e}_{ij,t} = \big[\, o_{ij,t}^{(x)},\ o_{ij,t}^{(y)},\ d_{ij,t},\ \cos(\theta_{ij,t}),\ \sin(\theta_{ij,t})\,\big] \in \mathbb{R}^5 $

This representation has three useful properties for failure detection: (i) relative edge encoding makes pairwise relations invariant to global translations; (ii) explicit modeling of pairwise object interactions, allowing failures involving object identity or spatial arrangement to be represented directly; and (iii) temporal consistency through velocity features and graph sequences, enabling detection of both instantaneous and sequential deviations.

\subsection{TPC: Geometric and Relational Consistency}
\label{subsec:tpc_method}

Our first scorer is non-parametric: it compares the observed episode directly to the set of expert trajectories, without fitting any model. For each tracked object, TPC computes the orthogonal projection distance from the observed trajectory to the closest expert trajectory of the same label, aggregated over time with exponentially increasing weights so that recent deviations dominate. Since the per-point distance does not depend on the prefix, the score of a streaming episode is maintained incrementally at constant cost per timestep. To capture deviations in interaction structure rather than in individual kinematics, TPC adds a correlation penalty: the $\ell_1$ divergence between the
pairwise Pearson correlation matrices of the observed objects and of the nominal dataset. Implementation details are given in Appendix~\ref{app:TPC_details}.

\begin{equation}
    S_{\text{total}}(t) = \sum_{i=1}^{C} \sum_{k=1}^{K(t)} \frac{e^{\gamma k}}{\sum_{j \le K(t)} e^{\gamma j}}
    \cdot \frac{\bigl| \min_{m} d\bigl(p_{i,k},\, \tau^{(m)}_{\ell(i)}\bigr) - \mu_{\ell(i)} \bigr|}{\sigma_{\ell(i)} + \varepsilon}
    \;+\; \sum_{i < j} \bigl| \rho_{ij}(t) - \bar{\rho}_{ij} \bigr|
\end{equation}

\subsection{GnnT: Predictive Spatio-Temporal Dynamics}
\label{subsec:prediction_method}
Our second scorer replaces the direct comparison to expert trajectories by a learned model of the nominal dynamics: failure detection is cast as a future-prediction problem, where deviations from predicted object states signal an anomaly. This formulation can represent action-reaction delays
that a purely geometric comparison cannot, at the cost of requiring enough demonstrations to fit the model.

\paragraph{Spatial Encoding via Dynamic Graph Attention}
Standard convolutional operators treat all neighbors equally. Instead, we employ GATv2~\cite{brody2022GATV2} to compute dynamic attention coefficients $\alpha_{ij,t}$, allowing the network to weight task-relevant interactions (e.g., an end-effector approaching a target object) more strongly than distant or weakly coupled entities. For a vertex $i$ and its neighbors $j \in \mathcal{N}_i$, the updated spatial embedding $\mathbf{h}_{i,t} \in \mathbb{R}^D$ is:
\begin{equation}
    \mathbf{h}_{i,t} = \sigma\left(\sum_{j \in \mathcal{N}_i} \alpha_{ij,t} \mathbf{W} \mathbf{x}_{j,t}\right)
\end{equation}
where $\mathbf{W}$ is a learnable weight matrix applied to the input features, and $\sigma$ denotes a non-linear activation function. This operation yields a set of spatially contextualized embeddings $H_t = \{\mathbf{h}_{1,t}, \dots, \mathbf{h}_{C,t}\}$, where $C$ is the total number of vertices in the graph.

\paragraph{Spatio-Temporal Transformer \& Future Prediction}
To model action-reaction delays across time, we process a sliding window of the past $m$ spatial embeddings $\mathcal{S}_{in} = \{H_{t-m+1}, \dots, H_t\}$. We flatten $\mathcal{S}_{in}$ into a sequence of $m \times C$ tokens. This fine-grained tokenization enables modeling dependencies between specific objects at specific past timesteps. We inject structural awareness via a composite positional encoding (PE) that accounts for relative time $\tau$ within the $m$-window, object ID $i$, and global episode progress:
\begin{equation}
    \mathbf{z}_{i,\tau}^{(0)} = \mathbf{h}_{i,t-(m-1)+\tau} + \text{PE}_{\text{time}}(\tau) + \text{PE}_{\text{obj}}(i) + \text{PE}_{\text{global}}\left(\frac{t}{T_{\max}}\right)
\end{equation}
After processing through $N$ Transformer encoder layers, the tokens corresponding to the current step $t$ are projected via an MLP to predict the future sequence of positions over horizon $L$: $\widehat{\mathbf{Y}}_{i,t} = \text{MLP}(\widehat{\mathbf{h}}_{i,t}) \in \mathbb{R}^{L \times d_{out}}$.

\paragraph{Training Objective}
The network is optimized solely on nominal demonstrations using a dual-loss objective. We minimize the Mean Squared Error (MSE) of the predicted trajectories ($\mathcal{L}_{\text{pos}}$) and add an edge-consistency term ($\mathcal{L}_{\text{edge}}$) encouraging agreement between predicted and observed pairwise spatial relations:
$
    \mathcal{L} = \lambda_p \mathcal{L}_{\text{pos}} + \lambda_e \mathcal{L}_{\text{edge}}
$.

\subsection{Uncertainty Calibration via Conformal Prediction}

A fundamental challenge in failure detection is defining the decision threshold without relying on arbitrary, task-specific manual tuning or exposing the model to OOD failure data. We address this using Conformal Prediction (CP)~\cite{lei2017distributionfreepredictiveinferenceregression, diquigiovanni2021importancebandfinitesampleexact}. Using a held-out calibration set of nominal episodes $\mathcal{Z}_N$, we compute the non-conformity scores (i.e., the prediction error for GnnT, or the geometric distance for TPC). For a user-specified tolerance level $\alpha \in (0,1)$, we build a time-varying threshold $\eta_t$ from these calibration scores with the functional construction of~\cite{diquigiovanni2021importancebandfinitesampleexact} (Appendix~\ref{app:cp}): its width is the $(1-\alpha)$-quantile of the maximum deviation of calibration episodes from the mean nominal profile. Under exchangeability, the corresponding finite-sample conformal construction would control the trajectory-level error rate; because we use the uncorrected empirical quantile, we treat the resulting level as approximate.
\section{Evaluation}

\subsection{Dataset}

We evaluate Robot-WATCHDOG on two complementary datasets: (i) a public benchmark from prior work and (ii) a new dataset designed to stress-test relational and dynamic failure detection.

\paragraph{BotFails.}
We evaluate on BotFails~\cite{rolland2026failure}, a recent benchmark for robotic failure detection collected with LeRobot~\cite{cadene2024lerobot}. The full dataset spans \textbf{10 manipulation tasks} across domestic and industrial settings with diverse annotated anomaly types. To align with our object-centric formulation, we retain \textbf{4 tasks} involving rigid and distinguishable objects: \textbf{Table-setting}, \textbf{Dish storing}, \textbf{Vegetable sorting}, and \textbf{Groceries sorting}. Tasks involving deformable or fluid interactions are excluded, as they violate the discrete object abstraction underlying our method. Additional details are provided in Appendix~\ref{sec:datasets_appendix}.

\paragraph{Ours.}
To complement BotFails, we introduce a new dataset comprising two manipulation tasks with increased relational and dynamic complexity. The first task is a \textbf{table-setting} scenario requiring structured object placement and ordering. The second is a \textbf{waste-sorting} task on a conveyor belt, where objects exhibit continuous motion, leading to variability in initial states and interaction timing. This dataset serves two purposes: (i) providing a larger number of demonstrations for training data-driven models, and (ii) evaluating robustness under increased variability and less constrained dynamics. Further details are provided in Appendix~\ref{sec:datasets_appendix}.

\subsection{Baselines}

We compare Robot-WATCHDOG against a diverse set of anomaly detection methods spanning three complementary categories.

\textbf{Trajectory matching.}
These methods detect kinematic and geometric deviations by comparing observed trajectories to expert demonstrations. To ensure invariance to execution speed and temporal misalignment, AP-RMS and AP-Fréchet rely on arc-length resampling and evaluate prefix-based distances using Root Mean Square error and the discrete Fréchet distance~\cite{eiter1994computing}, respectively. Soft-DTW instead computes a global alignment cost using a differentiable soft-minimum formulation~\cite{Soft-DTW}, making it more robust to local temporal distortions.

\textbf{Learned representations.}
We evaluate density-based and reconstruction-based approaches trained exclusively on nominal data. Specifically, we consider Continuous Normalizing Flow models (logpZO, lopO)~\cite{xu2025FailDetect}, which estimate the likelihood of observations either in latent space or via exact log-likelihood computation, and an AutoEncoder (AE-Recon)~\cite{Shi_2021AE, hasan2016learningtemporalregularityvideo, Wang2018}, which uses reconstruction error as an anomaly score.

\textbf{Visual-semantic alignment.}
We include FIDeL~\cite{rolland2026failure}, a recent two-stage method that combines patch-level feature alignment via optimal transport with a vision-language model to filter semantically irrelevant anomalies.

\textbf{Ours.}
We compare against both variants of our framework: (i) \textbf{TPC}, a non-parametric method based on trajectory projection and relational correlation, and (ii) \textbf{GnnT}, a predictive spatio-temporal graph model that captures object interactions and temporal dependencies. Additional implementation details are provided in the Appendix \ref{sec:baselines}.







\subsection{Evaluation Protocol}

We evaluate Robot-WATCHDOG at the step level in two settings: Figure \ref{fig:raw_scoring} reports threshold-independent anomaly scoring, reporting AUPR (favored under strong class imbalance) and maximum MCC (a balanced summary over all confusion matrix terms), while Figure \ref{fig:E2E_results} reports end-to-end detection after Conformal Prediction, reporting calibrated MCC (TPR and TNR per task are given in Appendix~\ref{app:e2e_per_task}), capturing respectively failure interception and robustness to benign variability. Every number is averaged over \textbf{ten seeds}, a seed redrawing the expert train/calibration partition, the network initialisation and the conformal split, so the reported deviations measure sensitivity to which demonstrations a method was given. Calibrated results use the modulation $s_1$ of Appendix~\ref{app:cp} (inspired by \cite{xu2025FailDetect}) and are averaged over $\alpha\in[0.80,0.95]$.

\subsection{Results and discussion}

\begin{figure}
    \centering
    \includegraphics[width=\textwidth]{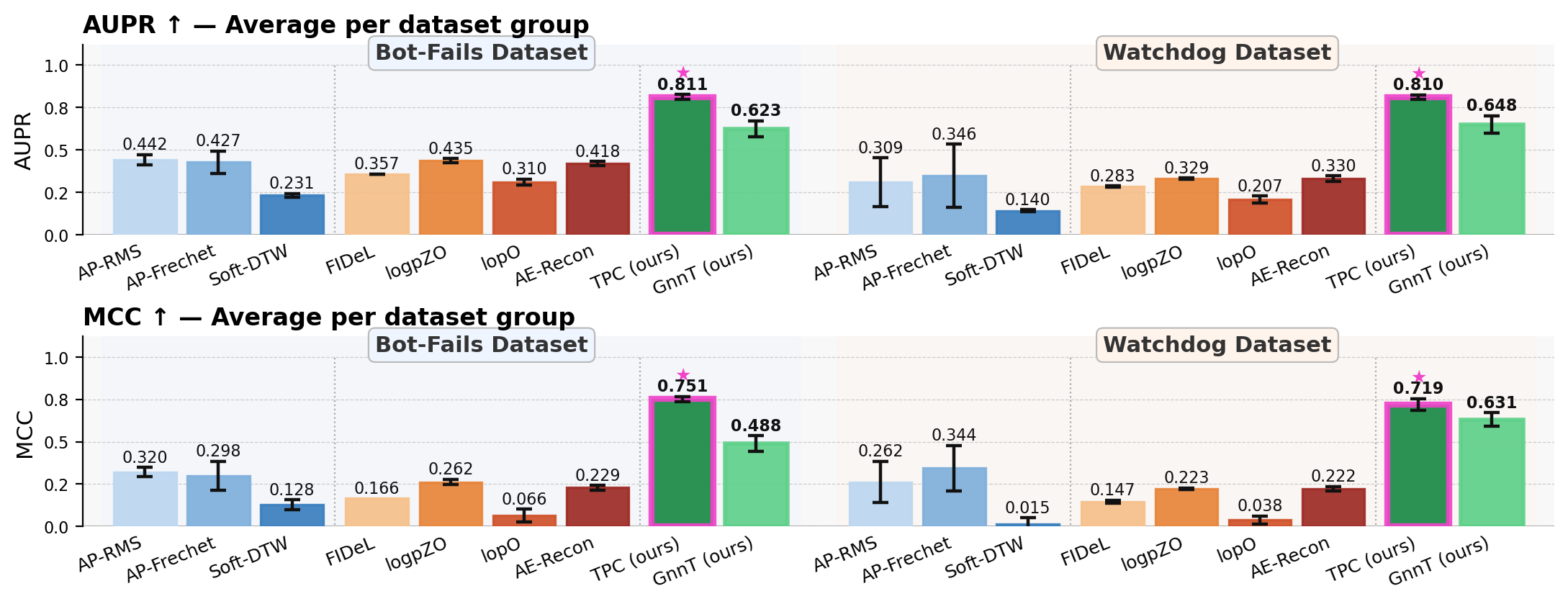}
    \caption{
        \textbf{Threshold-independent discriminative performance.} Average Area Under the Precision-Recall Curve (AUPR) and maximum Matthews Correlation Coefficient (MCC) across the BotFails and Watchdog datasets. Error bars denote the standard deviation across 10 seeds. Best results framed in purple with a star.
    }
    \label{fig:raw_scoring}
\end{figure}

\begin{figure}
    \centering
    \includegraphics[width=\textwidth]{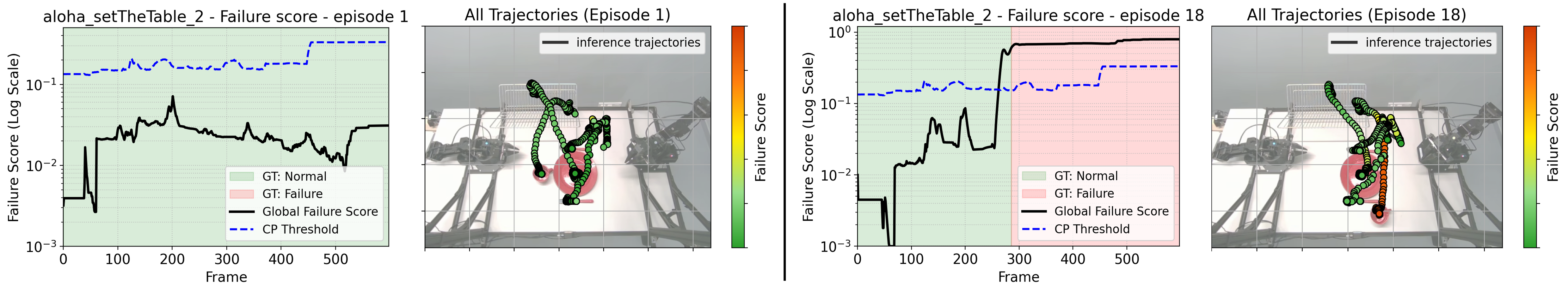}
    \caption{
        \textbf{Qualitative failure detection on the Table-setting task}. (Left) Nominal execution (Ep. 1): The global score remains safely below the CP threshold. (Right) Relational failure (Ep. 18): The cup is spatially misplaced. Due to natural spatial variability, absolute kinematics cannot detect this relational error. However, our object-centric graph captures the relational positional inconsistency, triggering a global score spike above the threshold that aligns with the misplaced cup's elevated local anomaly score (red trajectory points).
    }
    \label{fig:quali_fig}
\end{figure}

Figure~\ref{fig:raw_scoring} summarizes threshold-independent performance: TPC leads on both datasets (AUPR 0.811 / MCC 0.751 on BotFails, 0.810 / 0.719 on Watchdog), ahead of GnnT (0.623 / 0.488 and 0.648 / 0.631) and of the strongest baselines, AP-RMS on BotFails (0.442 / 0.320) and AP-Fr\'echet on Watchdog (0.346 / 0.344). Trajectory metrics evaluate objects independently, cannot express relational deviations, and are the least stable of all, with seed-to-seed standard deviations up to 0.19 AUPR on the Watchdog average (0.37 on its table-setting task), an order of magnitude above TPC (0.013); density, reconstruction and visual baselines (logpZO, lopO, AE-Recon, FIDeL) are steadier but plateau well below both of our scores. Figure~\ref{fig:quali_fig} illustrates a relational failure (misplaced cup) flagged TPC although absolute kinematics remain nominal.

\subsubsection{End-to-End Failure Detection}

\begin{wrapfigure}{r}{0.48\textwidth}
    \centering
    \vspace{-10pt}
    \includegraphics[width=\linewidth]{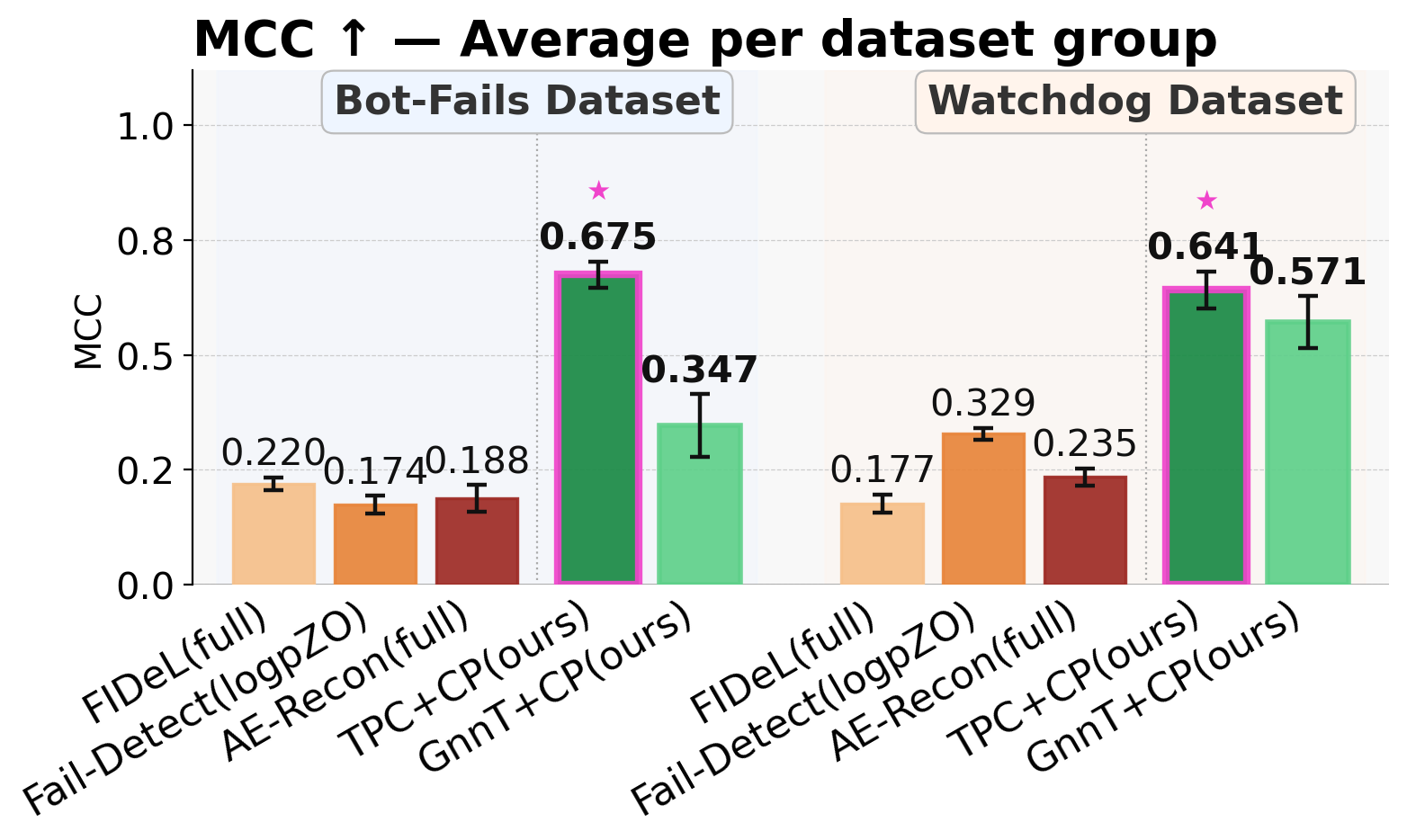}
    \vspace{-10pt}
    \caption{
    End-to-end failure detection performance (MCC) after conformal calibration (modulation $s_1$), averaged over the tasks of each dataset, over 10 seeds and over $\alpha\in[0.80,0.95]$. Error bars: standard deviation across seeds. 
    }
    \label{fig:E2E_results}
    \vspace{-10pt}
\end{wrapfigure}

With CP converting scores into binary alerts (Fig.~\ref{fig:E2E_results}), TPC again leads on both datasets ($0.675$ on BotFails, $0.641$ on Watchdog) ahead of GnnT ($0.347$, $0.571$), the best online baselines reaching $0.220$ (FIDeL, BotFails) and $0.329$ (logpZO, Watchdog). Gains in raw scoring do not transfer mechanically: baseline rankings change between the two settings, and each method has its own useful range, TPC peaking at low tolerance ($\alpha = 0.05$--$0.15$, up to MCC $0.845$ on Aloha table-setting) and GnnT only becoming competitive at $\alpha = 0.85$--$0.99$. The nominal level is not matched empirically either: $\alpha = 0.05$ flags $0.2$--$3\%$ of nominal steps, $\alpha = 0.99$ flags $10$--$15\%$. Calibration is thus a design choice to report in its own right, hence our two regimes and full sweep.

\section{Conclusion}
We introduced Robot-WATCHDOG, a self-supervised framework for robotic failure detection based on object-centric spatio-temporal graphs. On this representation, we studied two failure scores: TPC, a non-parametric measure of geometric and relational deviation from expert demonstrations, and GnnT, a predictive graph-transformer. Across a public benchmark and a newly collected dataset, both outperform image- and density-based detectors, particularly on failures involving object interactions and task structure. TPC is the strongest score overall, in both data regimes we tested, while requiring no training and the lowest inference cost; GnnT only approaches it in the continuously dynamic conveyor task, which indicates where learned dynamics may justify their additional complexity. More broadly, these results highlight object-centric relational representations as a promising direction for scalable real-time monitoring of learned robotic policies.


\section{Limitations}

Robot-WATCHDOG relies on a discrete object-centric abstraction, which limits applicability to tasks involving highly deformable materials, fluids, or interaction domains without clear object decomposition (e.g., painting or soldering). Performance also depends on the upstream vision-language tracking pipeline: severe occlusions, missed detections, or identity switches can degrade graph quality and propagate to the anomaly detector. Finally, our current formulation models object centroids only, and therefore cannot capture failures involving orientation or fine-grained pose. Extending the graph representation with 6D pose estimates is a natural direction for future work.


\clearpage
\acknowledgments{The authors would like to thank Charlotte Sczwarc for the magnificent illustration of the dog figure \ref{fig:intro_fig}, and Saad Aziz Alaoui for his valuable insights as well as Melanie Cassedanne for her unconditional support. 
This publication was made possible by the use of the CEA List FactoryAI supercomputer, financially supported by the Ile-de-France Regional Council. This work was partly funded by : the European Union’s Horizon Europe Research and Innovation program under grant agreement n\textsuperscript{o} 101135708 (JARVIS project) and n\textsuperscript{o} 101070596 (euROBIN project) - and by the France 2030 program through the PEPR O2R projects AS3 (ANR-22-EXOD-007).}


\bibliography{example}  

\newpage

\appendix
\section*{\centering \Large Appendix}
\addcontentsline{toc}{section}{Appendix}

\startcontents[appendix]
\printcontents[appendix]{l}{1}{\setcounter{tocdepth}{2}}
\newpage

\section{Ablation Study}
\label{app:ablations}

We conduct a series of ablation experiments to quantify the contribution of each component of the proposed framework. Our analysis focuses on (i) the input representation, (ii) relational modeling, (iii) the spatio-temporal architecture, and (iv) the predictive formulation.

\subsection{Impact of Object-Centric Graph Representation}

We evaluate whether representing the scene as an object-centric graph provides a measurable advantage over alternative representations.

In Table~\ref{tab:ablation_representation_wide}, we compare three configurations:

\paragraph{Vision feature baseline.}
To isolate the contribution of our Spatio-Temporal Graph Neural Network (GnnT) and its object-centric inductive bias, we introduce a purely sequential vision-based baseline that bypasses the graph construction phase. In this ablation, the demonstration episode is modeled as a flat sequence of global observations. At each timestep $t$, dense visual features extracted by a pre-trained encoder $\mathcal{F}_{\theta}$ (ResNet18 \cite{he2016deep}) are concatenated with proprioceptive data to form a state vector $\mathbf{x}_t \in \mathbb{R}^{d_{in}}$. These states are linearly projected into a hidden dimension $D$, augmented with sinusoidal positional encodings, and processed by a multi-layer Transformer Encoder to capture global temporal dependencies across the sequence. The resulting output tokens are then mapped back to the original feature space via an MLP head to reconstruct the input states $\mathbf{x}_t$. Trained exclusively on nominal data using a Mean Squared Error (MSE) objective, this auto-encoding baseline utilizes the state-wise reconstruction error as its failure score during inference. Results are presented in Table~\ref{tab:ablation_representation_wide} under the name \textit{Vision features}.

\paragraph{Trajectory-only representation.} We evaluate a graph-free ablation model, denoted as the Trajectory Transformer. This baseline deliberately removes the Graph Neural Network (GNN) encoder and all explicit edge-feature computations. Specifically, for each object $c \in \{1, \dots, C\}$ at time step $t \in \{1, \dots, m\}$, its kinematic state $x_c^t \in \mathbb{R}^2$ is concatenated with a learned semantic embedding $e_c \in \mathbb{R}^{d_{sem}}$ and linearly projected to a hidden token representation: $h_c^t = \mathbf{W} [x_c^t \parallel e_c] + \mathbf{b}$, where $\mathbf{W} \in \mathbb{R}^{d_h \times (2 + d_{sem})}$ is the projection weight matrix and $\parallel$ denotes concatenation. Following the addition of standard temporal positional encodings, the flattened sequence of independent object tokens is processed by a multi-head Transformer encoder. The future trajectories $\hat{Y} \in \mathbb{R}^{L \times C \times 2}$ are regressed directly from the final sequence embeddings via a Multi-Layer Perceptron (MLP) head. Unlike the full model, this baseline is optimized exclusively using a Mean Squared Error (MSE) loss on absolute coordinates, without any topological constraints. By stripping away structural priors, this ablation forces the self-attention mechanism to independently discover and model inter-object physical correlations solely from raw trajectories.

\paragraph{Graph-based representation (ours).} Object trajectories are structured as spatio-temporal graphs and processed by the proposed GNN-Transformer architecture; see Section~\textbf{3-Method} of the paper for more details.

\begin{table*}[h]
\centering
\small
\setlength{\tabcolsep}{3pt}
\resizebox{\textwidth}{!}{
\begin{tabular}{lcccccccc}
\toprule
& \multicolumn{4}{c}{\textbf{Waste-sorting}}
& \multicolumn{4}{c}{\textbf{Table-setting}} \\
\cmidrule(lr){2-5} \cmidrule(lr){6-9}
\textbf{Representation}
& TPR $\uparrow$ & TNR $\uparrow$ & AUPR $\uparrow$ & MCC $\uparrow$
& TPR $\uparrow$ & TNR $\uparrow$ & AUPR $\uparrow$ & MCC $\uparrow$ \\
\midrule

Vision features
& $29.7\,\pm\,0.2$ & $84.3\,\pm\,1.7$ & $29.7\,\pm\,2.4$ & $16.7\,\pm\,1.8$
& $30.6\,\pm\,0.6$ & $0.0\,\pm\,0.0$ & $23.3\,\pm\,1.2$ & $0.0\,\pm\,0.0$ \\

Trajectories
& $58.7\,\pm\,1.2$ & $\mathbf{92.9}\,\pm\,0.8$ & $\mathbf{54.7}\,\pm\,2.5$ & $\mathbf{56.5}\,\pm\,1.9$
& $54.5\,\pm\,1.6$ & $90.9\,\pm\,1.2$ & $54.5\,\pm\,1.7$ & $49.3\,\pm\,1.7$ \\

\rowcolor{lightgreen}
\textbf{Graph (Ours)}
& $\mathbf{81.8}\,\pm\,5.3$ & $79.9\,\pm\,3.5$ & $49.7\,\pm\,4.0$ & $53.1\,\pm\,1.9$
& $\mathbf{83.6}\,\pm\,6.9$ & $\mathbf{92.4}\,\pm\,3.6$ & $\mathbf{80.0}\,\pm\,7.3$ & $\mathbf{73.1}\,\pm\,6.6$ \\

\bottomrule
\end{tabular}
}
\caption{
Quantitative evaluation of different input representations on failure detection performance. We compare our proposed object-centric graph representation against a global vision feature baseline and a graph-free trajectory-only model across the WATCHDOG dataset. Metrics reported include True Positive Rate (TPR), True Negative Rate (TNR), Area Under the Precision-Recall Curve (AUPR), and Matthews Correlation Coefficient (MCC). All values are reported as percentages, with higher values ($\uparrow$) indicating better performance. The best results for each metric are highlighted in bold.
}
\label{tab:ablation_representation_wide}
\end{table*}

The results in Table~\ref{tab:ablation_representation_wide} highlight the importance of structured, object-level inputs. The global vision feature baseline performs poorly on both tasks and fails entirely on \textit{Table-setting} (TNR and MCC of $0$), indicating that unstructured, dense visual features lack the precise spatial inductive biases required for failure detection. Extracting explicit object trajectories yields a large improvement over raw vision features. The additional benefit of the graph structure, however, is strongly task-dependent. On \textit{Table-setting}, where failures are defined by incorrect spatial arrangements between objects, the graph representation outperforms the trajectory-only model on every metric, improving TPR by $29.1$ points, AUPR by $25.5$ points and MCC by $23.8$ points. On \textit{Waste-sorting}, the picture is more nuanced: the graph representation substantially increases sensitivity (TPR $+23.1$ points) at the cost of more false alarms (TNR $-13.0$ points), and the trajectory-only model retains a slightly higher AUPR ($54.7\%$ vs.\ $49.7\%$) and MCC ($56.5\%$ vs.\ $53.1\%$). This is consistent with the nature of \textit{Waste-sorting} failures, which mostly involve deviations of individual objects that can already be detected from their own trajectories, leaving limited room for relational reasoning. Overall, explicit relational structure is most valuable precisely in tasks whose failures are defined by inter-object configurations.

\subsection{Impact of Relational Modeling}

We evaluate the importance of explicitly modeling interactions between objects for both the predictive (GnnT) and non-parametric (TPC) formulations.

\paragraph{For the GnnT method}

To assess the contribution of relational inductive bias, we construct a randomized graph baseline by perturbing the edge structure while preserving node features. Formally, given a graph $\mathcal{G} = (\mathcal{V}, \mathcal{E})$ with $|\mathcal{E}| = E$, we sample a randomized edge set $\tilde{\mathcal{E}}$ by drawing $E$ pairs $(i,j)$ uniformly from $\mathcal{V} \times \mathcal{V}$, yielding a graph $\tilde{\mathcal{G}} = (\mathcal{V}, \tilde{\mathcal{E}})$. The node representations are then computed as
\[
\mathbf{h}_i^{(l+1)} = \sigma \left( \sum_{j \in \mathcal{N}_{\tilde{\mathcal{G}}}(i)} \alpha_{ij}^{(l)} \mathbf{W}^{(l)} \mathbf{h}_j^{(l)} \right),
\]
where $\alpha_{ij}^{(l)}$ denotes attention coefficients as in GATv2~\cite{brody2022GATV2}. This procedure preserves the computational structure of the model while destroying meaningful spatial correlations, enabling a controlled evaluation of the role of relational structure.

\paragraph{For the TPC method}

To disentangle the contribution of geometric versus relational cues in TPC, we evaluate three variants of the anomaly score:

\begin{itemize}
    \item \textbf{Orthogonal projection only.} Each object trajectory is compared independently to expert demonstrations using the weighted orthogonal projection distance. This captures local kinematic deviations but ignores interactions between objects.

    \item \textbf{Correlation only.} We discard absolute trajectories and instead compare the pairwise correlation structure between object motions. Concretely, for each pair of objects, we compute the Pearson correlation of their $x$ and $y$ trajectories, and penalize deviations from the correlation patterns observed in expert demonstrations. This captures relational consistency but is invariant to global geometric shifts.

    \item \textbf{Full TPC (projection + correlation).} The final score combines both terms, aggregating per-object projection errors with a global correlation penalty. This jointly captures absolute trajectory deviations and interaction-level inconsistencies.
\end{itemize}

\begin{table*}[h]
\centering
\small
\setlength{\tabcolsep}{3pt}
\resizebox{\textwidth}{!}{
\begin{tabular}{lcccccccc}
\toprule
& \multicolumn{4}{c}{\textbf{Waste-sorting}}
& \multicolumn{4}{c}{\textbf{Table-setting}} \\
\cmidrule(lr){2-5} \cmidrule(lr){6-9}
\textbf{Method}
& TPR $\uparrow$ & TNR $\uparrow$ & AUPR $\uparrow$ & MCC $\uparrow$
& TPR $\uparrow$ & TNR $\uparrow$ & AUPR $\uparrow$ & MCC $\uparrow$ \\
\midrule

\rowcolor{lightgreen}
\textbf{GnnT}
& $\mathbf{81.8}\,\pm\,5.3$ & $79.9\,\pm\,3.5$ & $\mathbf{49.7}\,\pm\,4.0$ & $\mathbf{53.1}\,\pm\,1.9$
& $\mathbf{83.6}\,\pm\,6.9$ & $\mathbf{92.4}\,\pm\,3.6$ & $\mathbf{80.0}\,\pm\,7.3$ & $\mathbf{73.1}\,\pm\,6.6$ \\

w/o Edge model
& $73.7\,\pm\,5.6$ & $\mathbf{80.9}\,\pm\,8.3$ & $48.5\,\pm\,5.2$ & $49.0\,\pm\,6.3$
& $81.8\,\pm\,6.9$ & $86.8\,\pm\,1.0$ & $59.3\,\pm\,9.3$ & $61.9\,\pm\,6.3$ \\

\midrule

Orthogonal projection only
& $\mathbf{89.2}\,\pm\,0$ & $87.3\,\pm\,0$ & $\mathbf{67.7}\,\pm\,0$ & $\mathbf{61.0}\,\pm\,0$
& $83.7\,\pm\,0$ & $95.1\,\pm\,0$ & $92.5\,\pm\,0$ & $78.2\,\pm\,0$ \\

Correlation only
& $82.1\,\pm\,0$ & $86.1\,\pm\,0$ & $65.1\,\pm\,0$ & $53.9\,\pm\,0$
& $\mathbf{94.9}\,\pm\,0$ & $91.2\,\pm\,0$ & $90.8\,\pm\,0$ & $81.3\,\pm\,0$ \\

\rowcolor{lightgreen}
Both \textbf{(TPC)}
& $47.7\,\pm\,1.9$ & $\mathbf{96.8}\,\pm\,1.2$ & $65.8\,\pm\,1.5$ & $55.1\,\pm\,2.7$
& $92.3\,\pm\,5.4$ & $\mathbf{97.1}\,\pm\,2.1$ & $\mathbf{96.3}\,\pm\,2.3$ & $\mathbf{88.8}\,\pm\,5.7$ \\

\bottomrule
\end{tabular}
}
\caption{
Ablation study on relational modeling for both GnnT and TPC. Best results are highlighted in bold within each block. For GnnT, removing edge-based interactions degrades performance, most strongly on \textit{Table-setting}, confirming that structured object relationships are important for robust failure detection. For TPC, geometric (projection) and relational (correlation) components capture complementary aspects of anomalies: projection alone is sufficient for trajectory-level deviations in \textit{Waste-sorting}, while correlation improves the detection of interaction-level errors in \textit{Table-setting}. Combining both signals yields the best AUPR, TNR and MCC on \textit{Table-setting}, whereas on \textit{Waste-sorting} it leads to a conservative operating point (highest TNR, lower TPR), highlighting the task-dependent role of relational modeling.
}
\label{tab:ablation_relations_wide}
\end{table*}

\paragraph{Results and Interpretation}

The results in Table~\ref{tab:ablation_relations_wide} highlight distinct roles of geometric and relational signals in failure detection, as well as their task-dependent interplay.

For GnnT, removing the edge model degrades performance on both tasks, but the magnitude of the drop depends on the relational complexity of the task. On \textit{Table-setting}, the degradation is severe on threshold-independent and balanced metrics (AUPR $-20.7$ points, MCC $-11.2$ points) and is accompanied by more false alarms (TNR $-5.6$ points), showing that the model cannot reliably distinguish benign variations from relational failures without explicit interactions. On \textit{Waste-sorting}, the effect is concentrated on sensitivity (TPR $-8.1$ points, MCC $-4.1$ points), while TNR and AUPR remain comparable. Removing the edge model also makes the predictions noticeably less stable across seeds (e.g., TNR standard deviation of $8.3$ vs.\ $3.5$ on \textit{Waste-sorting}), indicating that relational structure also acts as a regularizer.

For TPC, the orthogonal projection term alone achieves the best TPR, AUPR and MCC on \textit{Waste-sorting}, suggesting that failures in this task are predominantly characterized by deviations from nominal trajectories. In contrast, the correlation-only variant attains the highest TPR on \textit{Table-setting}, indicating that many failures arise from incorrect coordination between objects despite locally plausible motions. Combining both signals yields the best overall performance on \textit{Table-setting}, with the highest TNR ($97.1\%$), AUPR ($96.3\%$) and MCC ($88.8\%$, i.e.\ $+7.5$ points over correlation only and $+10.6$ points over projection only), while keeping a TPR close to the correlation-only variant ($92.3\%$ vs.\ $94.9\%$). This confirms that geometric deviations and relational inconsistencies provide complementary information. On \textit{Waste-sorting}, however, the combined score shifts towards a conservative operating point: it achieves the highest TNR ($96.8\%$) but a markedly lower TPR ($47.7\%$), resulting in a lower MCC than projection only ($55.1\%$ vs.\ $61.0\%$). Since its AUPR remains close to that of projection only ($65.8\%$ vs.\ $67.7\%$), the ranking ability of the score is largely preserved, and the drop in sensitivity mainly reflects the decision threshold rather than a loss of discriminative power. This suggests that, in simpler tasks dominated by spatial accuracy, the relational term brings limited additional information. Overall, these results indicate that effective failure detection requires capturing both absolute trajectory deviations and interaction-level structure, with their relative importance depending on the semantic complexity of the task.

\subsection{Impact of Spatio-Temporal Modeling}

To rigorously evaluate the contribution of our combined GNN–Transformer architecture, we conduct an ablation study isolating the spatial and temporal inductive biases. We compare the full model against three degraded baselines:

\begin{itemize}
    \item \textbf{Full model (GnnT):} Our proposed architecture, which uses an Edge-aware Graph Attention Network (EGAT) to explicitly model spatial interactions and physical constraints between objects, followed by a Transformer encoder to capture long-term temporal dependencies.

    \item \textbf{Transformer-only (Temporal without Spatial Priors):} The graph encoder is completely removed. Raw object coordinates and semantic embeddings are linearly projected into a hidden space and fed directly to the Transformer. This baseline tests whether the self-attention mechanism alone is sufficient to implicitly discover inter-object physical correlations without the structural priors provided by the GNN.

    \item \textbf{GNN-only (Spatial without Temporal Attention):} The temporal sequence modeling is replaced by a static feed-forward network. Spatial embeddings are still extracted frame-by-frame via the GNN, but they are subsequently concatenated along the temporal axis and mapped to future predictions via a Multi-Layer Perceptron (MLP). This isolates the value of the Transformer's sequence-awareness.

    \item \textbf{MLP (No explicit priors):} A naive baseline where both the GNN and the Transformer are removed. The kinematic data of the scene over the observation window is flattened and processed by a simple MLP. This serves as a lower bound, representing a purely data-driven regression without structural or sequential inductive biases.
\end{itemize}

\begin{table*}[h]
\centering
\small
\setlength{\tabcolsep}{3pt}
\caption{
    \textbf{Quantitative results of the spatio-temporal architecture ablation.}
    Metrics are reported as mean $\pm$ standard deviation over multiple random seeds. The full GnnT architecture achieves the best Matthews Correlation Coefficient (MCC) on both tasks and the best Area Under the Precision-Recall curve (AUPR) on \textit{Table-setting}, providing the best balance between detection sensitivity and false positive rejection. On \textit{Waste-sorting}, differences between the non-MLP variants are small.
}
\label{tab:ablation_architecture}
\resizebox{\textwidth}{!}{
\begin{tabular}{lcccccccc}
\toprule
& \multicolumn{4}{c}{\textbf{Waste-sorting}}
& \multicolumn{4}{c}{\textbf{Table-setting}} \\
\cmidrule(lr){2-5} \cmidrule(lr){6-9}
\textbf{Method}
& TPR $\uparrow$ & TNR $\uparrow$ & AUPR $\uparrow$ & MCC $\uparrow$
& TPR $\uparrow$ & TNR $\uparrow$ & AUPR $\uparrow$ & MCC $\uparrow$ \\
\midrule

\rowcolor{lightgreen}
\textbf{GnnT}
& $81.8\,\pm\,5.3$ & $79.9\,\pm\,3.5$ & $49.7\,\pm\,4.0$ & $\mathbf{53.1}\,\pm\,1.9$
& $83.6\,\pm\,6.9$ & $92.4\,\pm\,3.6$ & $\mathbf{80.0}\,\pm\,7.3$ & $\mathbf{73.1}\,\pm\,6.6$ \\

Transformer only
& $\mathbf{82.3}\,\pm\,6.0$ & $78.8\,\pm\,3.9$ & $50.8\,\pm\,3.7$ & $52.2\,\pm\,3.0$
& $87.0\,\pm\,7.0$ & $87.2\,\pm\,2.8$ & $66.2\,\pm\,8.1$ & $66.6\,\pm\,8.3$ \\

GNN only
& $70.8\,\pm\,5.6$ & $\mathbf{82.9}\,\pm\,5.3$ & $\mathbf{51.8}\,\pm\,1.4$ & $48.6\,\pm\,4.0$
& $83.3\,\pm\,3.0$ & $\mathbf{92.8}\,\pm\,2.8$ & $78.1\,\pm\,6.0$ & $71.2\,\pm\,4.2$ \\

MLP
& $75.3\,\pm\,12.8$ & $70.6\,\pm\,13.1$ & $40.5\,\pm\,2.9$ & $39.3\,\pm\,3.6$
& $\mathbf{88.7}\,\pm\,6.6$ & $46.2\,\pm\,7.3$ & $31.4\,\pm\,3.7$ & $28.8\,\pm\,5.1$ \\

\bottomrule
\end{tabular}
}
\end{table*}

\paragraph{Results and Discussion}
The empirical results, summarized in Table~\ref{tab:ablation_architecture}, show that the naive MLP baseline yields the lowest AUPR and MCC on both tasks. Its high TPR on \textit{Table-setting} ($88.7\%$) comes at the cost of a very low TNR ($46.2\%$), meaning that it flags a large fraction of nominal episodes as failures. Conversely, the full GnnT architecture achieves the highest MCC on both tasks and the highest AUPR on \textit{Table-setting}. This gap validates our hypothesis that explicitly modeling spatio-temporal priors is essential for robust failure detection in multi-object manipulation.

The ablation further reveals that the benefit of each inductive bias depends on the task's primary failure modes. In the highly structured \textit{Table-setting} task, anomalies frequently stem from incorrect geometric configurations or collisions. Without explicit relational modeling, the Transformer-only baseline struggles to capture these spatial constraints and clearly underperforms the GNN-only model (AUPR of $66.2\%$ vs.\ $78.1\%$, MCC of $66.6\%$ vs.\ $71.2\%$), and adding temporal attention on top of the GNN further improves both metrics ($80.0\%$ AUPR and $73.1\%$ MCC for GnnT). In the \textit{Waste-sorting} task, where the topological structure is less critical and anomalies often manifest as kinematic or timing deviations (e.g., missing an object), the differences between the non-MLP variants are much smaller and largely within one standard deviation. The sequence-aware Transformer-only model reaches a TPR and MCC on par with GnnT ($82.3\%$ / $52.2\%$ vs.\ $81.8\%$ / $53.1\%$), whereas the temporally blind GNN-only model trades sensitivity for specificity (TPR of $70.8\%$, TNR of $82.9\%$) and obtains a lower MCC ($48.6\%$). By combining both components, GnnT obtains the best MCC on both tasks, with the lowest variance on \textit{Waste-sorting} ($\pm 1.9$), making it the most reliable configuration across spatial and temporal failure regimes.

\subsection{Prediction vs Density Estimation}

We compare the predictive formulation with a likelihood-based alternative. Results are presented in Table~\ref{tab:ablation_model_density}.

\begin{itemize}
    \item \textbf{GnnT, Prediction-based (ours).} The model predicts future trajectories and uses prediction error as the failure score.

    \item \textbf{Density-based.} A normalizing flow is trained on latent representations to estimate likelihood.
\end{itemize}

\subsubsection{Latent Density Estimation via Normalizing Flows}
Instead of directly predicting the future trajectories of objects, we formulate the failure detection task as estimating the exact likelihood of the observed spatio-temporal dynamics. Let $\mathbf{h}_t \in \mathbb{R}^D$ be the latent representation of the scene at time $t$, extracted by the Spatio-Temporal Transformer from the observation window. The distribution of nominal interactions $p_H(\mathbf{h}_t)$ is highly complex and intractable.

To model this density, we employ a Normalizing Flow (NF) parameterized by $\theta$. The flow applies a sequence of invertible, differentiable transformations $f_\theta$ to map the complex latent space $\mathbf{h}_t$ into a simple, tractable base distribution, typically an isotropic multivariate Gaussian $\mathbf{z}_t \sim \mathcal{N}(\mathbf{0}, \mathbf{I})$.

Using the change of variables formula, the exact log-likelihood of a nominal representation is given by:
\begin{equation}
    \log p_H(\mathbf{h}_t) = \log p_Z(f_\theta(\mathbf{h}_t)) + \log \left| \det \left( \frac{\partial f_\theta(\mathbf{h}_t)}{\partial \mathbf{h}_t} \right) \right|
\end{equation}
where $p_Z$ is the standard Gaussian density function, and the second term is the log-determinant of the Jacobian of the transformation, accounting for the change in volume induced by $f_\theta$.

\paragraph{Training Objective.} The model is trained end-to-end to maximize the log-likelihood of the nominal data representations. Since $\mathbf{z}_t = f_\theta(\mathbf{h}_t)$ follows a standard normal distribution, maximizing $\log p_H(\mathbf{h}_t)$ is equivalent to minimizing the negative log-likelihood loss $\mathcal{L}_{\text{NF}}$:
\begin{equation}
    \mathcal{L}_{\text{NF}} = \frac{1}{2} \| f_\theta(\mathbf{h}_t) \|_2^2 - \log \left| \det \left( \frac{\partial f_\theta(\mathbf{h}_t)}{\partial \mathbf{h}_t} \right) \right|
\end{equation}

\paragraph{Inference and Failure Scoring.} During deployment, the Normalizing Flow acts as an explicit density estimator. For a new observation window, the model computes the latent representation $\mathbf{h}_t$ and its corresponding negative log-likelihood. The failure score $S_t$ is directly defined as $S_t = \mathcal{L}_{\text{NF}}(\mathbf{h}_t)$. A high score $S_t$ indicates that the observed spatio-temporal interaction pattern has a low probability under the learned nominal distribution.

\begin{table*}[h]
\centering
\small
\setlength{\tabcolsep}{3pt}
\resizebox{\textwidth}{!}{
\begin{tabular}{lcccccccc}
\toprule
& \multicolumn{4}{c}{\textbf{Waste-sorting}}
& \multicolumn{4}{c}{\textbf{Table-setting}} \\
\cmidrule(lr){2-5} \cmidrule(lr){6-9}
\textbf{Method}
& TPR $\uparrow$ & TNR $\uparrow$ & AUPR $\uparrow$ & MCC $\uparrow$
& TPR $\uparrow$ & TNR $\uparrow$ & AUPR $\uparrow$ & MCC $\uparrow$ \\
\midrule

\rowcolor{lightgreen}
\textbf{GnnT (prediction based)}
& $\mathbf{81.8}\,\pm\,5.3$ & $79.9\,\pm\,3.5$ & $\mathbf{49.7}\,\pm\,4.0$ & $\mathbf{53.1}\,\pm\,1.9$
& $83.6\,\pm\,6.9$ & $\mathbf{92.4}\,\pm\,3.6$ & $\mathbf{80.0}\,\pm\,7.3$ & $\mathbf{73.1}\,\pm\,6.6$ \\

Density based
& $70.2\,\pm\,6.7$ & $\mathbf{80.3}\,\pm\,4.0$ & $48.6\,\pm\,6.7$ & $45.1\,\pm\,3.7$
& $\mathbf{87.2}\,\pm\,3.5$ & $90.4\,\pm\,6.9$ & $78.2\,\pm\,11.7$ & $72.9\,\pm\,9.2$ \\

\bottomrule
\end{tabular}
}
\caption{
    Quantitative comparison of failure detection formulations: predictive modeling (GnnT) versus latent density estimation (Normalizing Flows). Mean and standard deviation are reported across 5 different seeds. The predictive approach achieves a substantially higher MCC on \textit{Waste-sorting} and matches or slightly exceeds the density-based baseline on \textit{Table-setting}, while exhibiting lower variance on the threshold-independent (AUPR) and balanced (MCC) metrics across both tasks.
}
\label{tab:ablation_model_density}
\end{table*}

As shown in Table~\ref{tab:ablation_model_density}, the predictive formulation (GnnT) matches or outperforms the density estimation baseline on the balanced metrics while being more stable. On \textit{Waste-sorting}, GnnT detects substantially more failures (TPR of $81.8\%$ vs.\ $70.2\%$) at an equivalent false positive rate (TNR of $79.9\%$ vs.\ $80.3\%$), which translates into an $8.0$-point gain in MCC. On \textit{Table-setting}, both formulations reach comparable AUPR ($80.0\%$ vs.\ $78.2\%$) and MCC ($73.1\%$ vs.\ $72.9\%$), but operate at different points: the density-based model is slightly more sensitive (TPR of $87.2\%$ vs.\ $83.6\%$), while GnnT produces fewer false alarms (TNR of $92.4\%$ vs.\ $90.4\%$). The most marked difference lies in stability: the density-based model exhibits considerably higher variance across seeds, reaching $\pm 11.7$ in AUPR and $\pm 9.2$ in MCC on \textit{Table-setting}, compared to $\pm 7.3$ and $\pm 6.6$ for GnnT. This suggests that fitting a normalizing flow to high-dimensional, complex spatio-temporal latents yields less reliable decision boundaries, whose quality strongly depends on the training run. Conversely, framing failure detection as future trajectory prediction enforces a strict structural prior on nominal behavior. By directly computing errors in the geometric observation space rather than relying on abstract latent likelihoods, the predictive formulation provides a more consistent separation between genuine failures and benign variations across tasks and seeds.

\section{Method Details}

\subsection{Trajectory Extraction}
\label{app:traj_extraction}

Object trajectories are extracted from RGB observations using a two-stage pipeline
combining vision-language-based object detection and segmentation-based multi-object
tracking. The pipeline is designed to be fully automated, requiring only a natural
language description of the task to identify and track all task-relevant entities
throughout an episode.

\paragraph{Stage 1: Object detection via VLM (Qwen3-VL)}
The first frame of each episode is passed to a vision-language model
(Qwen3-VL-8B-Instruct~\cite{bai2025qwen3vltechnicalreport}) together with a
task-specific prompt that enumerates the expected task-relevant objects.
For each object, the model is asked to predict a bounding box in normalised
coordinates $[0, 1000]^2$, a fine-grained semantic label, and a coarse-grained
label.
To improve robustness to occasional parsing failures, the model is queried up to
seven times with a temperature annealing schedule
($T_k = 0.10 + (k-1) \times 0.15$, $k \in \{1,\ldots,7\}$).
A response is accepted as valid only if all expected objects are returned with
non-degenerate bounding boxes and non-empty labels.
If the maximum number of attempts is exceeded, missing detections are flagged. This initialisation step is performed \textit{offline} prior to deployment and therefore imposes no latency on real-time execution. The prompts used for each task are provided below.

\begin{figure}[h]
    \centering
    \includegraphics[width=\textwidth]{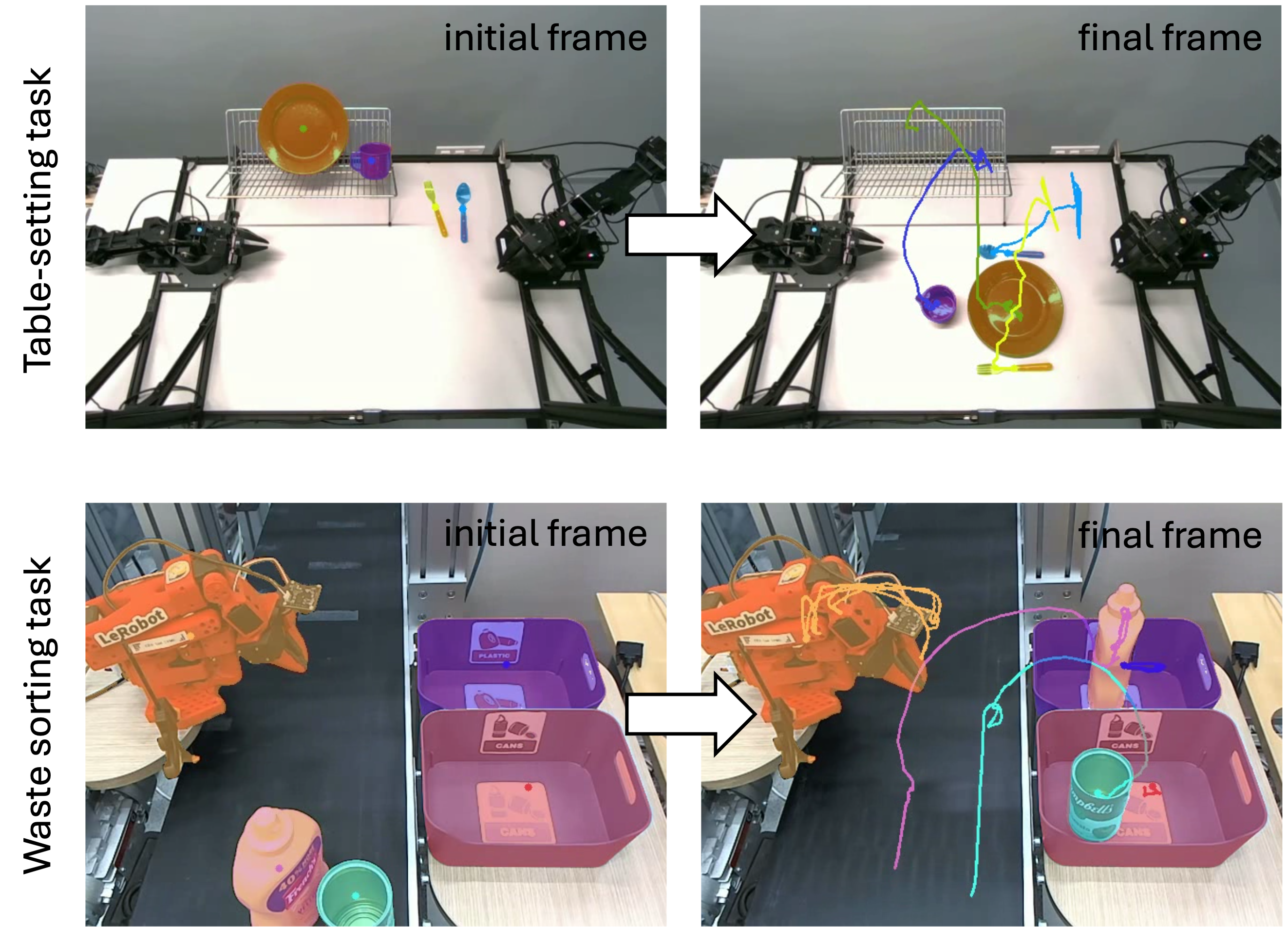}
    \caption{\textbf{Trajectory extraction pipeline.} (Left) The initial frame of an episode where task-relevant objects and the robot end-effector are identified using a Vision-Language Model (Stage 1). (Right) The final frame displaying the continuous 2D spatial trajectories extracted by tracking the objects' mask centroids throughout the episode via SAM2 (Stage 2). The top row illustrates the structured \textit{Table-setting} task, while the bottom row demonstrates the dynamic \textit{Waste sorting} task on a moving conveyor belt.}
    \label{fig:traj_extraction}
\end{figure}

\paragraph{Stage 2: Multi-object tracking via SAM2}
The bounding boxes produced in Stage 1 initialise a SAM2 video tracking session~\cite{ravi2024sam2}. At each subsequent frame, the segmentation masks are propagated using the SAM2 recurrent memory mechanism, and the 2D centroid of each object's mask is recorded as its spatial state $\mathbf{p}_{i,t} \in \mathbb{R}^2$. When the mask of an object degenerates (empty support), the last valid centroid is carried forward (\textit{forward-fill} imputation) so that all trajectories remain temporally aligned, while a validity label is switch from 1 to 0 to notify the missing object. If the object reappears later, the trajectory is interpolated between the last known position and the new position. The robot end-effector is treated as an additional object and tracked through the same pipeline from an initial bounding box provided by the VLM. Illustration of an example of the process is presented figure \ref{fig:traj_extraction}.

\paragraph{Real-time deployment characteristics}
The VLM initialisation is performed entirely offline; once bounding boxes are obtained, the real-time tracking pipeline relies solely on SAM2. In its optimised streaming configuration, SAM2 operates at approximately \SI{44} fps on an NVIDIA A100 GPU with 5 tracked objects, well above the \SI{30} fps acquisition rate of the camera used in our experiments. The resulting end-to-end perception latency is therefore bounded by the single-frame forward pass of SAM2 and does not accumulate over an episode.


\paragraph{Task-specific VLM prompts}
For reproducibility, we report the exact prompts supplied to Qwen3-VL for each task in our benchmark. Each prompt follows the same schema: a brief scene description, the user task string, a numbered list of the expected objects with their semantic categories, and an instruction to return a strictly formatted JSON response.

Table~\ref{tab:vlm_prompts} summarises the object lists used per task.
Full prompt strings are reproduced verbatim in the supplementary material.

\begin{table}[h]
\centering
\caption{Object lists supplied to Qwen3-VL per task.}
\label{tab:vlm_prompts}
\begin{tabular}{lll}
\toprule
Task & $N_\text{obj}$ & Object list \\
\midrule
\texttt{domotic\_groceriesSorting}         & 8 & 4 sweets, 4 vegetables \\
\texttt{domotic\_vegetablesAndFruitsSorting} & 6 & 3 fruits, 3 vegetables \\
\texttt{domotic\_dishTidyUp}               & 8 & 1 whisk, 3 plates, 3 glasses, 1 cup \\
\texttt{domotic\_setTheTable}              & 4 & 1 glass, 1 plate, 1 fork, 1 knife \\
\texttt{aloha\_setTheTable}              & 4 & 1 cup, 1 plate, 1 fork, 1 spoon \\
\texttt{sort\_trash\_conveyor}             & 2 & 1 plastic waste, 1 metal can \\

\bottomrule
\end{tabular}
\end{table}

All prompts share the following preamble and JSON schema:

\begin{tcolorbox}[title=Common preamble and output schema, fontupper=\small\ttfamily, breakable]
A robot arm is executing a task.\\
User task: "<TASK\_DESCRIPTION>"\\
Please identify this <N> objects in the image and provide their bounding box\\
as (x1, y1, x2, y2) in image coordinates.\\
The <N> objects:\\
\phantom{xx}- [object list -- see below per task]\\
Return your answer strictly in this JSON format and NOTHING ELSE,\\
make sure to keep objects in the correct order:\\
\\
\{\\
\phantom{xx}"objects": [\\
\phantom{xxxx}\{\\
\phantom{xxxxxx}"fine\_label":   "...",\\
\phantom{xxxxxx}"coarse\_label": "...",\\
\phantom{xxxxxx}"bbox":         [x1, y1, x2, y2],\\
\phantom{xxxxxx}"is\_related":   true\\
\phantom{xxxx}\}\\
\phantom{xx}]\\
\}\\
\\
Ensure the JSON is valid and contains no extra text or explanations.
\end{tcolorbox}

\bigskip

\begin{tcolorbox}[title=\texttt{domotic\_groceriesSorting} (8 objects), fontupper=\small\ttfamily, breakable]
A robot arm is executing a task.\\
User task: "<TASK\_DESCRIPTION>"\\
Please identify this 8 objects in the image and provide their bounding box\\
as (x1, y1, x2, y2) in image coordinates.\\
The 8 objects:\\
\phantom{xx}- 1. sweet.\\
\phantom{xx}- 2. sweet.\\
\phantom{xx}- 3. sweet.\\
\phantom{xx}- 4. sweet.\\
\phantom{xx}- 5. vegetal.\\
\phantom{xx}- 6. vegetal.\\
\phantom{xx}- 7. vegetal.\\
\phantom{xx}- 8. vegetal.\\
Return your answer strictly in this JSON format and NOTHING ELSE,\\
make sure to keep objects in the correct order.
\end{tcolorbox}

\begin{tcolorbox}[title=\texttt{domotic\_vegetablesAndFruitsSorting} (6 objects), fontupper=\small\ttfamily, breakable]
A robot arm is executing a task.\\
User task: "<TASK\_DESCRIPTION>"\\
Please identify this 6 objects in the image and provide their bounding box\\
as (x1, y1, x2, y2) in image coordinates.\\
The 6 objects:\\
\phantom{xx}- 1. fruit.\\
\phantom{xx}- 2. fruit.\\
\phantom{xx}- 3. fruit.\\
\phantom{xx}- 4. vegetable.\\
\phantom{xx}- 5. vegetable.\\
\phantom{xx}- 6. vegetable.\\
Return your answer strictly in this JSON format and NOTHING ELSE,\\
make sure to keep objects in the correct order.
\end{tcolorbox}

\begin{tcolorbox}[title=\texttt{domotic\_dishTidyUp} (8 objects), fontupper=\small\ttfamily, breakable]
A robot arm is executing a task.\\
User task: "<TASK\_DESCRIPTION>"\\
Please identify this 8 objects in the image and provide their bounding box\\
as (x1, y1, x2, y2) in image coordinates.\\
The 8 objects:\\
\phantom{xx}- 1. whisk.\\
\phantom{xx}- 2. plate.\\
\phantom{xx}- 3. plate.\\
\phantom{xx}- 4. plate.\\
\phantom{xx}- 5. glass.\\
\phantom{xx}- 6. glass.\\
\phantom{xx}- 7. glass.\\
\phantom{xx}- 8. cup.\\
Return your answer strictly in this JSON format and NOTHING ELSE,\\
make sure to keep objects in the correct order.
\end{tcolorbox}

\begin{tcolorbox}[title=\texttt{domotic\_setTheTable} / \texttt{aloha\_setTheTable\_2} (4 objects), fontupper=\small\ttfamily, breakable]
A robot arm is executing a task.\\
User task: "<TASK\_DESCRIPTION>"\\
Please identify this 4 objects in the image and provide their bounding box\\
as (x1, y1, x2, y2) in image coordinates.\\
The 4 objects:\\
\phantom{xx}- 1. glass.\\
\phantom{xx}- 2. plate.\\
\phantom{xx}- 3. fork.\\
\phantom{xx}- 4. knife.\\
Return your answer strictly in this JSON format and NOTHING ELSE,\\
make sure to keep objects in the correct order.
\end{tcolorbox}

\begin{tcolorbox}[title=\texttt{sort\_trash\_conveyor} (2 objects), fontupper=\small\ttfamily, breakable]
A robot arm is executing a task.\\
User task: "<TASK\_DESCRIPTION>"\\
Please identify this 2 objects in the image and provide their bounding box\\
as (x1, y1, x2, y2) in image coordinates.\\
The 2 objects:\\
\phantom{xx}- 1. a plastic waste -- fine\_label=plastic.\\
\phantom{xx}- 2. a can, metal waste -- fine\_label=metal.\\
Return your answer strictly in this JSON format and NOTHING ELSE,\\
make sure to keep objects in the correct order.
\end{tcolorbox}

\subsection{Trajectory Extraction Robustness}

\subsubsection{VLM Initialization Robustness}
The first step of our pipeline relies on a Vision-Language Model (VLM) to perform open-vocabulary object detection and establish the initial bounding boxes for subsequent tracking. To evaluate its robustness, we tested the Qwen-VL model on our dataset (341 episodes) both with and without our standardization loop.

As shown in Table~\ref{tab:vlm_robustness}, the base Qwen model achieves a correct detection rate of 92.4\%. However, it occasionally misses some of the objects (5.3\%) and suffers from identity confusion (1.8\%) with a wrong label attribution, and localization inaccuracies (0.6\%). By introducing our standardization verification loop (\textit{Qwen\_WithStd}), which re-prompts the model if the expected number of objects is not found, we attenuate the missing objects rate of 2.1\%, raising the overall success rate to 94.4\%.

\begin{table}[h]
\centering
\caption{VLM Initialization Robustness (evaluated on 341 episodes). "\textit{Missing}" indicates the model failed to detect all of the objects related to the task in the scene. "\textit{Identity}" means some of the label attributed to the objects are incorrect. \textit{Position} indicates a highly inaccurate bounding box.}
\label{tab:vlm_robustness}
\begin{tabular}{lcccc}
\toprule
\textbf{Method} & \textbf{Correct (\%)} & \textbf{Missing (\%)} & \textbf{Identity Error (\%)} & \textbf{Position Error (\%)} \\
\midrule
Qwen (No Stdz) & 92.4 & 5.3 & 1.8 & 0.6 \\
Qwen (With Stdz) & \textbf{94.4} & \textbf{3.2} & 1.8 & 0.6 \\
\bottomrule
\end{tabular}
\end{table}

\subsubsection{Tracking Robustness}
To ensure the VLM errors do not impact the tracking evaluation, we evaluated the SAM2 tracker's robustness using ground-truth initial bounding boxes. We categorize the tracking performance into four levels: \textit{Perfect} (object never lost), \textit{Recovered} (object temporarily lost due to occlusion but recovered before the end of the task), \textit{Missing} (object definitively lost by the tracker), and \textit{Fusion} (tracker confused the object with another nearby object). 
\begin{table}[h]
\centering
\caption{SAM2 Tracking Robustness initialized with Ground Truth bounding boxes.}
\label{tab:tracking_robustness}
\resizebox{\linewidth}{!}{
\begin{tabular}{lcccc}
\toprule
\textbf{Task Type} & \textbf{Perfect Tracking (\%)} & \textbf{Recovered (\%)} & \textbf{Missing (\%)} & \textbf{Fusion (\%)} \\
\midrule
\multicolumn{5}{c}{\textbf{Robot-Watchdog Datasets}} \\
\midrule
Aloha SetTheTable & 97.9 & 0.5 & 0.5 & 1.0 \\
Trash Conveyor & 97.7 & 0.2 & 0.9 & 1.1 \\
\textbf{Average (Robot-Watchdog)} & \textbf{97.8} & \textbf{0.3} & \textbf{0.8} & \textbf{1.1} \\
\midrule
\multicolumn{5}{c}{\textbf{Botfails Datasets}} \\
\midrule
Domotic Sorting (Fruits \& Veg) & 98.3 & 0.0 & 1.7 & 0.0 \\
Domotic SetTheTable & 98.5 & 0.0 & 1.5 & 0.0 \\
Domotic Groceries Sorting & 98.8 & 0.0 & 1.2 & 0.0 \\
Domotic Dish Tidy Up & 99.4 & 0.0 & 0.6 & 0.0 \\
\textbf{Average (Botfails)} & \textbf{98.8} & \textbf{0.0} & \textbf{1.2} & \textbf{0.0} \\
\midrule
\textbf{Global Average} & \textbf{98.3} & \textbf{0.2} & \textbf{1.0} & \textbf{0.6} \\
\bottomrule
\end{tabular}
}
\end{table}

\subsection{GnnT - Model Architecture}

The spatial encoder is implemented using a Graph Attention Network (GATv2), while temporal dependencies are modeled using a multi-layer Transformer encoder. The prediction head consists of a shared MLP applied independently to each object embedding.

\paragraph{Background}
Spatio-temporal forecasting in multi-agent or multi-object scenarios requires simultaneously resolving relational structures and temporal dynamics. While standard Graph Neural Networks (GNNs) effectively capture spatial topologies, they lack mechanisms for long-term sequence modeling. Conversely, Transformers excel at sequential processing but lack structural priors for geometric arrangements. The \textit{GnnT} (Graph Neural Network - Transformer) architecture marries these paradigms, utilizing an edge-aware spatial encoder to independently process intra-frame geometry, followed by a sequence model to capture inter-frame kinematics.

\paragraph{Hypothesis}
We hypothesize that nominal robotic manipulation is governed by strict, predictable physical laws and interaction patterns (e.g., collisions, grasping semantics, kinematic constraints). By explicitly forcing a network to predict future states from an observation window, it should internalize these underlying rules. The dynamic attention mechanism of the GATv2 is expected to learn to attend heavily to interacting entities (e.g., the end-effector approaching a target), while the Transformer learns temporal delays, inertia, and task phases. Consequently, mechanical or procedural failures will intrinsically violate these learned spatio-temporal priors, manifesting as a sharp spike in the prediction error ($\mathcal{L}_{\text{pos}}$ and $\mathcal{L}_{\text{edge}}$), which serves as a highly discriminative anomaly signal.

\paragraph{Implementation details}
The model takes as input a sequence of $m$ graphs containing $C$ objects. Let the raw kinematic state of object $i$ at time $t$ be $\mathbf{x}_{i,t} \in \mathbb{R}^2$. To provide semantic identity, each object's discrete category $c_i \in \{1, \dots, C\}$ is mapped to a continuous embedding $\mathbf{e}_i \in \mathbb{R}^{16}$. During training, we inject Gaussian noise $\epsilon \sim \mathcal{N}(0, \sigma^2 \mathbf{I})$ with $\sigma = 0.01$ to the coordinates to prevent overfitting and encourage the learning of robust physical dynamics rather than trivial identity mappings. The initial node feature vector is thus the concatenation:
$$ \mathbf{v}_{i,t}^{(0)} = [\,(\mathbf{x}_{i,t} + \epsilon) \parallel \mathbf{e}_i \,] \in \mathbb{R}^{18} $$

The spatial encoder employs a two-layer Edge-aware GATv2 (EGAT). Let $\mathbf{V}_t^{(l)}$ and $\mathbf{E}_t$ denote the set of node and edge features at time $t$ for layer $l$. The spatial embedding is refined iteratively:
$$ \mathbf{v}_{i,t}^{(1)} = \text{ELU}\left( \text{EGAT}^{(1)}\big(\mathbf{v}_{i,t}^{(0)}, \mathbf{E}_t\big) \right), \quad \mathbf{v}_{i,t}^{(2)} = \text{ELU}\left( \text{EGAT}^{(2)}\big(\mathbf{v}_{i,t}^{(1)}, \mathbf{E}_t\big) \right) $$
where $\text{ELU}(\cdot)$ is the Exponential Linear Unit activation used to maintain non-linear representational capacity while preventing dying gradients, and $\mathbf{v}_{i,t}^{(2)} \in \mathbb{R}^{d_h}$ (with $d_h=128$) represents the spatially contextualized node embedding.

To model sequence dynamics, standard sinusoidal temporal positional encodings $\mathbf{p}_t \in \mathbb{R}^{d_h}$ are added to the spatial embeddings: $\tilde{\mathbf{h}}_{i,t} = \mathbf{v}_{i,t}^{(2)} + \mathbf{p}_t$. The sequence of $m$ spatial graphs is then flattened into a single contiguous token sequence:
$$ \mathcal{S}_{in} = \big[ \tilde{\mathbf{h}}_{1,t-m+1}, \dots, \tilde{\mathbf{h}}_{C,t-m+1}, \dots, \tilde{\mathbf{h}}_{1,t}, \dots, \tilde{\mathbf{h}}_{C,t} \big] \in \mathbb{R}^{mC \times d_h} $$
This flattened topology permits unrestricted, fine-grained self-attention across both the temporal and spatial axes. The sequence is processed by a $4$-layer Transformer Encoder to yield $\mathcal{S}_{out} = \text{Transformer}(\mathcal{S}_{in})$. 

Finally, we extract only the $C$ tokens corresponding to the most recent observation at timestep $t$ (i.e., the last $C$ elements of $\mathcal{S}_{out}$). Let $\mathbf{z}_{i,t} \in \mathbb{R}^{d_h}$ denote this output token for object $i$. The future trajectory over horizon $L$ is directly regressed via a two-layer Multi-Layer Perceptron (MLP) with a ReLU activation:
$$ \widehat{\mathbf{Y}}_{i,t} = \mathbf{W}_2 \max(0, \mathbf{W}_1 \mathbf{z}_{i,t} + \mathbf{b}_1) + \mathbf{b}_2 \quad \in \mathbb{R}^{L \times 2} $$
where $\mathbf{W}_1 \in \mathbb{R}^{2d_h \times d_h}$ and $\mathbf{W}_2 \in \mathbb{R}^{2L \times 2d_h}$ are the shared projection weights applied independently to each object token.

\paragraph{Hyperparameters}
The architecture and training routine are governed by the hyperparameters detailed in Table~\ref{tab:hyperparameters}. These values were selected empirically to balance the temporal receptive field, representational capacity, and computational efficiency.

\begin{table}[htpb]
    \centering
    \caption{Hyperparameters for the \textit{GnnT} architecture and training routine.}
    \label{tab:hyperparameters}
    \begin{tabular}{@{}llc@{}}
        \toprule
        \textbf{Category} & \textbf{Parameter} & \textbf{Value} \\
        \midrule
        \multirow{2}{*}{\textit{Sequence}} 
        & Observation window ($m$) & $20$ \\
        & Prediction horizon ($L$) & $15$ \\
        \midrule
        \multirow{6}{*}{\textit{Architecture}} 
        & Hidden dimension ($d_h$) & $128$ \\
        & Spatial Encoder layers (GATv2) & $2$ \\
        & Spatial Encoder attention heads & $8$ \\
        & Temporal Encoder layers (Transformer) & $4$ \\
        & Temporal Encoder attention heads & $8$ \\
        & Tokenization format & Batch-first \\
        \midrule
        \multirow{5}{*}{\textit{Optimization \& Training}} 
        & Optimizer & AdamW \\
        & Learning rate & $10^{-3}$ \\
        & Weight decay & $10^{-4}$ \\
        & Batch size & $256$ \\
        & Total epochs & $100$ \\
        \bottomrule
    \end{tabular}
\end{table}

\subsection{Trajectory Projection and Correlation (TPC) Formulation}
\label{app:TPC_details}

The TPC method evaluates kinematic and relational anomalies by comparing inference trajectories against a set of nominal expert demonstrations. Let a tracked object's trajectory of length $T$ be defined as $P = \{p_1, \dots, p_T\}$ with $p_t \in \mathbb{R}^2$, and let the set of nominal expert trajectories for the corresponding object class be $\mathcal{E} = \{V^{(1)}, \dots, V^{(K)}\}$. Prior to distance computation, trajectories are filtered to remove static segments, ensuring the evaluation focuses strictly on kinematic progression.

\paragraph{1. Projection-Based Kinematic Scoring:}
The spatial deviation of a point $p_t$ from an expert trajectory $V$ (modeled as a piecewise linear polyline) is defined by the orthogonal point-to-polyline distance, denoted $d(p_t, V)$. The spatial failure at time $t$ is taken as the distance to the nearest expert:
\begin{equation}
    d^*(p_t) = \min_{V \in \mathcal{E}} d(p_t, V)
\end{equation}
To account for the inherent natural variability of the task, these distances are standard-normalized using the empirical mean $\mu$ and standard deviation $\sigma$ of the pairwise projection distances computed within the expert set itself:
\begin{equation}
    \hat{d}(p_t) = \frac{|d^*(p_t) - \mu|}{\sigma + \epsilon}
\end{equation}
where $\epsilon$ is a small constant for numerical stability. The individual projection score aggregates these normalized deviations over time. To penalize errors that compound towards the end of the execution, temporal weights $w_t$ (e.g., derived from an exponential decay function) are applied:
\begin{equation}
    S_{\text{proj}} = \sum_{t=1}^{T} w_t \hat{d}(p_t)
\end{equation}

\paragraph{2. Relational Correlation Penalty:}
To capture higher-level interaction failures, the method evaluates the synchronized movement between objects. For any pair of objects $i$ and $j$ resampled to a uniform length $T$, let $\rho(x^i, x^j)$ and $\rho(y^i, y^j)$ be the Pearson correlation coefficients of their respective spatial coordinates over time. The pairwise spatial correlation is averaged across dimensions:
\begin{equation}
    C_{i,j} = \frac{1}{2}\big(\rho(x^i, x^j) + \rho(y^i, y^j)\big)
\end{equation}
The relational failure penalty is computed as the $\ell_1$ distance between the inference correlation matrix ($C^{\text{inf}}$) and the reference correlation matrix ($C^{\text{exp}}$) derived from the expert dataset:
\begin{equation}
    S_{\text{corr}} = \sum_{i=1}^{N} \sum_{j > i} \left| C^{\text{inf}}_{i,j} - C^{\text{exp}}_{i,j} \right|
\end{equation}
where $N$ is the number of tracked objects.

\paragraph{3. Final failure Score:}
Depending on the configuration, the method can evaluate purely projection-based errors, purely relational errors, or a combination of both. The combined failure score is the sum of the individual projection scores and the relational penalty:
\begin{equation}
    S_{\text{total}} = S_{\text{proj}} + S_{\text{corr}}
\end{equation}

\subsection{Conformal Prediction}
\label{app:cp}

\subsubsection{Calibration Data}

Calibrating the decision band requires a reference set of nominal trajectories, denoted
$\mathcal{Z}_{N} = \{\tau^{(1)}, \dots, \tau^{(N)}\}$. The expert set is partitioned once per seed: one half trains the detector (or, for the
training-free scorers, serves as the reference set of expert trajectories), and the complementary half forms $\mathcal{Z}_{N}$. No demonstration is used both to fit the detector and to calibrate its threshold.

At every timestep, the per-object non-conformity scores are summed across the tracked objects to yield a single scalar $S_t$, and non-finite values are mapped to zero before calibration.

\subsubsection{Sequential Band Construction}

Rather than a single global threshold, we calibrate a time-varying band, following the functional split-conformal construction of Diquigiovanni et al. \cite{diquigiovanni2021importancebandfinitesampleexact} The calibration set is split a second time into two disjoint subsets, $\mathcal{D}_A$ (a fraction $\rho$ of $\mathcal{Z}_N$) and $\mathcal{D}_B$ (the remainder). $\mathcal{D}_A$ estimates the shape of the band and $\mathcal{D}_B$ its width, so that the width is never fitted on the same data that defines the
centre.

From $\mathcal{D}_A$ we compute the pointwise mean score
$$\mu_t = \frac{1}{|\mathcal{D}_A|} \sum_{i \in \mathcal{D}_A} S_t^{(i)}$$
together with a modulation function $s(t)$ that sets how the band width is distributed over time.
We consider two variants:
$$s_{1}(t) = \frac{1}{T}, \qquad
  s_{2}(t) = \max_{i \in \mathcal{D}_A} \left| S_t^{(i)} - \mu_t \right| .$$
The constant modulation $s_1$ yields a band that tracks the mean expert profile with a uniform offset, whereas $s_2$ widens the band wherever the expert demonstrations themselves disagree.

The width is then calibrated on $\mathcal{D}_B$. For each trajectory $j \in \mathcal{D}_B$ we take the maximum deviation from the mean profile, normalised by the modulation function: $$h^{(j)} = \max_{t \in [1,T]} \frac{\left| \mu_t - S_t^{(j)} \right|}{s(t)} ,$$ and set $h$ to the empirical $(1-\alpha)$-quantile of $\{h^{(j)}\}_{j \in \mathcal{D}_B}$. Taking the maximum over the whole horizon before the quantile makes the coverage statement simultaneous over the rollout rather than pointwise at a single timestep. The resulting threshold is
$$\eta_t = \mu_t + h \cdot s(t) .$$

\subsubsection{Online Deployment and Evaluation}

At test time the framework computes the instantaneous score $S_t$ and applies the one-sided decision
rule $\mathbb{1}(S_t > \eta_t)$: only high scores indicate out-of-distribution behaviour, and a
failure is flagged at the earliest timestep where the band is breached. Test rollouts may outlast
the calibration demonstrations; beyond the calibration horizon the last band value $\eta_{T}$ is
held constant.

Because $h$ is the empirical quantile of the maximum normalised deviation over $\mathcal{D}_B$,
the fraction of nominal trajectories flagged at any point of their execution is controlled at
approximately $\alpha$ in the exchangeable regime. We use the plain empirical quantile rather than
the finite-sample corrected $\lceil (1-\alpha)(|\mathcal{D}_B|+1) \rceil$ order statistic, so the
guarantee is asymptotic; with the calibration set sizes used here ($|\mathcal{D}_B|$ of a few tens
of demonstrations) the distinction is not negligible and the nominal level should be read as
indicative.

We report $s_1$ throughout, and sweep $\alpha$ over $\{0.05, 0.10, \dots, 0.95, 0.99\}$ for both
modulation variants; headline numbers average the operating points $\alpha \in [0.80, 0.95]$, and
every $(\alpha, s)$ configuration is reported in the supplementary material. All results are
averaged over 10 seeds, each seed redrawing the expert partition, the $\mathcal{D}_A/\mathcal{D}_B$
split and the network initialisation.

\subsection{Pipeline Latency and Real-Time Capability}
A critical requirement for our Robot-Watchdog pipeline is the ability to operate in real-time alongside the robot's control loop. We profiled the latency of each component on an NVIDIA RTX 3090 GPU.
Table~\ref{tab:latency} details the execution time of each module. The VLM initialization is an expensive operation ($\sim$7.0s) but is only executed once at the beginning of the task. For the continuous monitoring loop, SAM2 segmentation achieves highly efficient real-time performance at approximately 22.72 ms per frame (44 FPS). On the anomaly detection side, the Graph Neural Network Transformer (GnnT) inference takes 19.87 ms per frame, while TPC sliding window inference is extremely lightweight, requiring only 7.00 ms. 
This low computational overhead enables the overall pipeline to operate in true real-time. When using TPC, the end-to-end processing time is 29.72 ms per frame (33.6 FPS), comfortably exceeding standard 30 FPS control loop requirements. When using the more complex GnnT, the pipeline runs at 23.5 FPS, which remains fully suitable for standard 15-20 FPS robotic control setups.

\begin{table}[h]
\centering
\caption{Latency Profiling of the Robot-Watchdog Pipeline components (NVIDIA RTX 3090).}
\label{tab:latency}
\begin{tabular}{lcc}
\toprule
\textbf{Component} & \textbf{Execution Time} & \textbf{Frequency / Nature} \\
\midrule
VLM Initialization (Qwen) & 7.03 s & Once per task \\
SAM2 Tracking & 22.72 ms & Per frame \\
GnnT Inference & 19.87 ms & Per frame \\
TPC Inference & 7.00 ms & Per frame \\
\midrule
\textbf{End-to-End (SAM2 + GnnT)} & \textbf{42.59 ms} & \textbf{23.5 FPS} \\
\textbf{End-to-End (SAM2 + TPC)} & \textbf{29.72 ms} & \textbf{33.6 FPS} \\
\bottomrule
\end{tabular}
\end{table}

\section{Experimental Details}

\subsection{Datasets}
\label{sec:datasets_appendix}

We evaluate Robot-WATCHDOG on two datasets: the BotFails benchmark~\cite{rolland2026failure} and a new dataset collected for this work. Together, they provide complementary coverage of controlled failure scenarios and more realistic, dynamically varying environments.

\paragraph{BotFails.}
BotFails is a recently introduced dataset for robotic failure detection, collected via teleoperation on a real robotic platform. It includes multimodal observations (vision, proprioception, and language) across 10 tasks spanning domestic and industrial settings, with annotations for both general anomalies and task-relevant failures.

In our evaluation, we select a subset of tasks compatible with our object-centric assumptions. Specifically, we retain tasks involving rigid, non-deformable, and visually distinguishable objects, enabling consistent tracking and relational reasoning. Tasks involving fluids or deformable materials (e.g., pouring) are excluded, as they violate the assumption of persistent object identity. 

\begin{figure}[h]
    \centering
    \includegraphics[width=\textwidth]{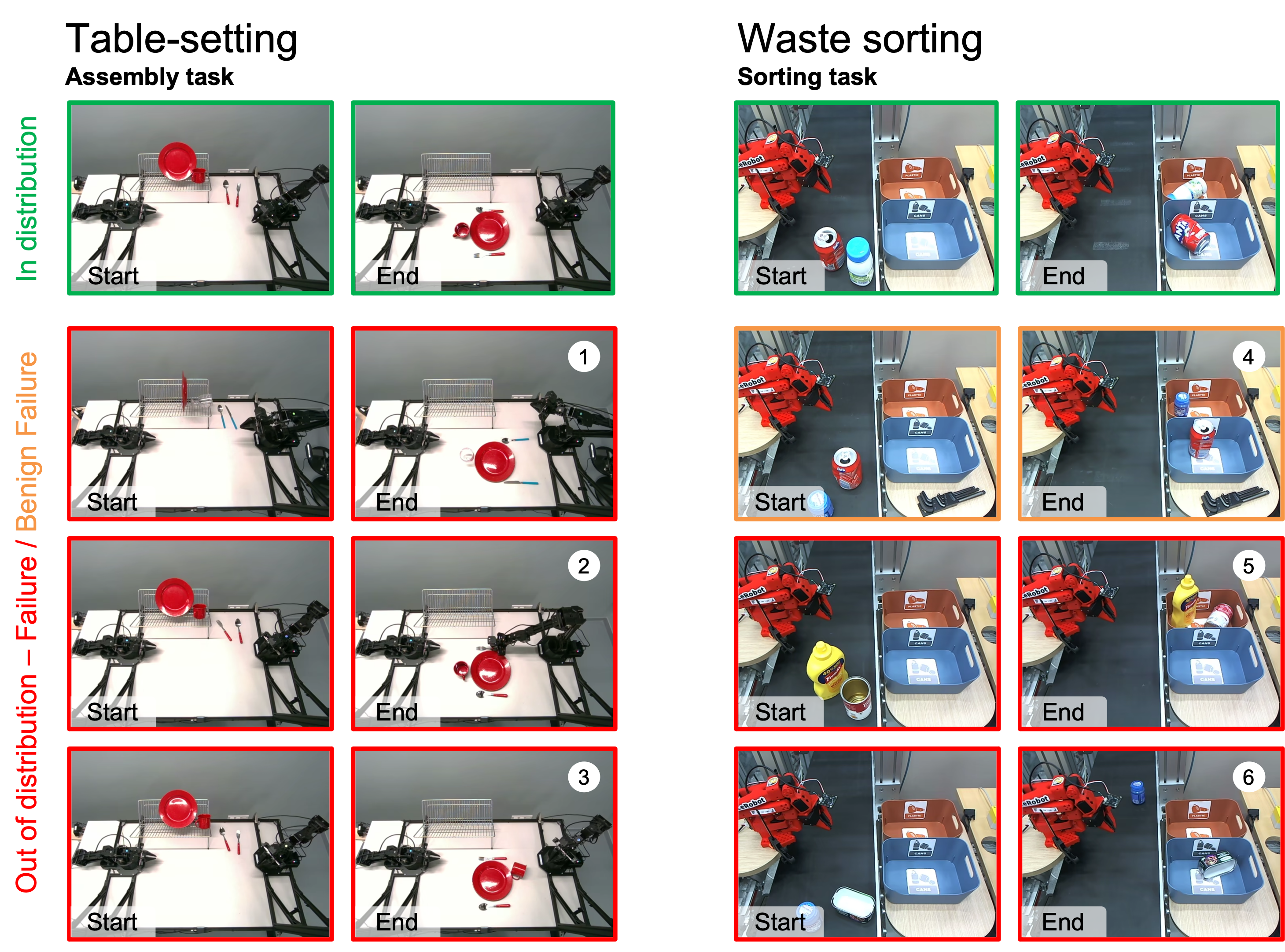}
    \caption{\textbf{Overview of the Robot-WATCHDOG evaluation dataset.} The dataset comprises a structured Table-setting task (left) and a dynamic Waste sorting task (right). The top row illustrates in-distribution, nominal expert demonstrations. The subsequent rows depict out-of-distribution scenarios, categorized into critical task failures (red borders) and benign visual variations (orange borders). Specific evaluation scenarios include: (1) Spatial misplacement, where objects are arranged with an incorrect offset; (2) Object inversion, where the target locations of items are swapped; (3) Grasping failure, resulting in a dropped object; (4) Benign visual variation featuring an intrusive distractor object. This highlights our method's robustness to visual clutter, which typically triggers false positives in baseline anomaly detectors; (5) identity sorting error, where an item is placed into the incorrect bin; and (6) Missed object, where the robot fails to make a decision or initiate a grasp.}
    \label{fig:watchdog_dataset}
\end{figure}

\paragraph{Ours.}
We introduce a new dataset, see figure \ref{fig:watchdog_dataset}, composed of two manipulation tasks designed to increase both relational and dynamic complexity:

\begin{itemize}
    \item \textbf{Table-setting:} a structured assembly task where the robot must place objects (e.g., plates, utensils) in predefined spatial configurations and correct temporal order. This task emphasizes relational correctness (object identity, placement location, ordering).
    
    \item \textbf{Waste sorting:} a dynamic task where the robot sorts objects on a moving conveyor belt. Unlike static tabletop setups, object positions evolve continuously, introducing variability in initial conditions and interaction timing.
\end{itemize}

For each task, we collect two types of datasets: (i) \textit{expert demonstrations}, used for training, and (ii) \textit{failure datasets}, used for evaluation and containing both nominal and failure trajectories.

\paragraph{Data collection and modalities.}
The table-setting task is performed using an ALOHA robot, while the waste-sorting task uses a SO-100 platform. Observations include RGB video streams and proprioceptive states (joint positions), along with corresponding action commands. The table-setting setup includes multiple camera viewpoints (external and wrist-mounted), whereas the sorting task uses a top-down camera.

\begin{table}[t]
\centering
\caption{Statistics of our dataset.}
\label{tab:dataset_stats}
\resizebox{\linewidth}{!}{
\begin{tabular}{lcccccccc}
\toprule
\textbf{Task} & \textbf{Split} & \textbf{Data} & \textbf{Robot} & \textbf{\# Episodes} & \textbf{\# Frames} & \textbf{FPS} & \textbf{Cameras} & \textbf{Action dim} \\
\midrule
Table-setting & Expert & Video RGB/Proprio/text & ALOHA & 100 & 67,341 & 15 & 4 views & 9\\
Table-setting & Test & Video RGB/Proprio/text & ALOHA & 21 & 13,471 & 15 & 4 views & 9 \\
\midrule
Waste sorting & Expert  & Video RGB/Proprio/text & SO-100 & 100 & 59,562 & 30 & Top view & 6\\
Waste sorting & Test  & Video RGB/Proprio/text & SO-100 & 51 & 30,378 & 30 & Top view & 6\\
\bottomrule
\end{tabular}
}
\end{table}

\paragraph{Discussion.}
Our dataset complements BotFails in two key aspects. First, it provides a larger number of expert demonstrations, which is beneficial for training predictive models of scene dynamics. Second, it introduces more challenging conditions for failure detection, particularly in the sorting task where object motion induces continuous distribution shift. This setup allows us to evaluate robustness to both relational or identity errors (e.g., wrong object or placement) and dynamic deviations (e.g., delayed or missed interactions).

\subsection{Baselines}
\label{sec:baselines}

We compare our methods against classical and learning-based failure detection methods, which characteristics are presented in table~\ref{tab:comparison_baselines}.

\begin{table*}[t]
\centering
\caption{Overview of the anomaly detection baselines compared in this work, including our proposed \textbf{TPC} (few-shot) and \textbf{GnnT} (predictive graph) approaches.}
\label{tab:comparison_baselines}
\resizebox{\textwidth}{!}{%
\begin{tabular}{l l l l l l} 
\toprule
\textbf{Method} & \textbf{Input Space} & \textbf{Core Mechanism} & \textbf{Spatial Rel.} & \textbf{Temporal Align.} & \textbf{Scoring Strategy} \\ 
\midrule
AP-Fréchet & Trajectories (2D) & Trajectory Matching & Independent & Arc-length Resampling & Min. Fréchet Distance \\
Soft-DTW & Trajectories (2D) & Trajectory Matching & Independent & Soft-DTW & Min. Alignment Cost \\
AP-RMS & Trajectories (2D) & Trajectory Matching & Independent & Arc-length Resampling & Min. Prefix RMS Error \\
\midrule
FIDeL \cite{rolland2026failure} & Patch Features (RGB) & Optimal Transport & Implicit (Patches) & Min. Cost over $T$ & OT Cost (+ VLM Filter) \\
logpZO \cite{xu2025FailDetect} & Observations & Density Estimation & Implicit (Network) & ODE Forward Pass & Latent Space $\ell_2$-norm \\
lopO \cite{xu2025FailDetect} & Observations & Density Estimation & Implicit (Network) & ODE Integration & Exact Log-likelihood \\
AE-Recon \cite{Shi_2021AE} & Observations & Reconstruction & Implicit (Network) & Frame-wise (None) & Reconstruction Error \\
\midrule
\rowcolor{gray!10}
\textbf{TPC (Ours)} & Trajectories (2D) & Trajectory Matching & \textbf{Explicit (Pearson)} & Fixed Resampling & Proj. Dist. + Corr. Penalty \\
\rowcolor{gray!10}
\textbf{GnnT (Ours)} & \textbf{Spatio-Temporal Graphs} & \textbf{Future Prediction} & \textbf{Explicit (GATv2)} & \textbf{Sliding Window (Attn)} & \textbf{Prediction Error} \\
\bottomrule
\end{tabular}%
}
\end{table*}

\subsubsection{Arc-Length Prefix RMS (AP-RMS)}
The AP-RMS baseline detects kinematic deviations by evaluating the cumulative spatial error of a trajectory's prefix relative to a set of nominal expert trajectories. To ensure the comparison is invariant to execution speed and sampling rates, all trajectories are spatially aligned via arc-length parameterization.

\paragraph{1. Arc-Length Resampling:}
Let a raw trajectory consisting of $N$ points be defined as $P = \{p_1, \dots, p_N\}$ with $p_i \in \mathbb{R}^2$. The cumulative arc-length $s_i$ at step $i$ is computed as:
\begin{equation}
    s_i = \sum_{j=2}^{i} \|p_j - p_{j-1}\|_2, \quad \text{with } s_1 = 0
\end{equation}
The sequence is parameterized by normalized arc-length $\bar{s}_i = s_i / s_N \in [0, 1]$. We construct a continuous trajectory function $f(\bar{s})$ via linear interpolation. The trajectory is then uniformly resampled into a fixed length $T$, producing the spatially aligned trajectory $\tilde{P} = \{\tilde{p}_1, \dots, \tilde{p}_T\}$, where $\tilde{p}_\tau = f\left(\frac{\tau-1}{T-1}\right)$. This procedure is applied identically to the set of expert trajectories to form the aligned expert set $\tilde{\mathcal{E}} = \{\tilde{V}^{(1)}, \dots, \tilde{V}^{(K)}\}$.

\paragraph{2. Prefix RMS Distance:}
To evaluate the deviation of the execution up to a given progression step $t \in [1, T]$, we extract the trajectory prefix $\tilde{P}_{1:t} = \{\tilde{p}_1, \dots, \tilde{p}_t\}$. The Root Mean Square (RMS) distance between the observed prefix and the corresponding prefix of an expert $\tilde{V}^{(k)}_{1:t}$ is defined as:
\begin{equation}
    d_{\text{RMS}}\big(\tilde{P}_{1:t}, \tilde{V}^{(k)}_{1:t}\big) = \sqrt{ \frac{1}{t} \sum_{\tau=1}^{t} \big\| \tilde{p}_\tau - \tilde{v}^{(k)}_\tau \big\|_2^2 }
\end{equation}

\paragraph{3. Final failure Score:}
At each progression step $t$, the spatial failure is quantified as the distance to the closest expert prefix. This local deviation is given by:
\begin{equation}
    d^*(t) = \min_{\tilde{V} \in \tilde{\mathcal{E}}} d_{\text{RMS}}\big(\tilde{P}_{1:t}, \tilde{V}_{1:t}\big)
\end{equation}
The overall failure score for the complete trajectory is the temporal average of these minimal prefix deviations, capturing the continuous divergence from the nominal behavior manifold:
\begin{equation}
    S_{\text{total}} = \frac{1}{T} \sum_{t=1}^{T} d^*(t)
\end{equation}

\subsubsection{Arc-Length Prefix Fréchet (AP-Fréchet) Formulation}

The AP-Fréchet baseline detects structural and geometric anomalies by evaluating the shape discrepancy of trajectory prefixes relative to nominal expert demonstrations. Similar to the AP-RMS method, this approach enforces temporal invariance by spatially aligning trajectories via arc-length parameterization.

\paragraph{1. Arc-Length Resampling:}
We apply the identical arc-length normalization procedure defined for AP-RMS. Let the uniformly resampled observed trajectory of length $T$ be $\tilde{P} = \{\tilde{p}_1, \dots, \tilde{p}_T\}$, and the corresponding aligned expert set be $\tilde{\mathcal{E}} = \{\tilde{V}^{(1)}, \dots, \tilde{V}^{(K)}\}$. 

\paragraph{2. Discrete Fréchet Distance:}
To evaluate the geometric divergence up to a progression step $t \in [1, T]$, we extract the trajectory prefixes $\tilde{P}_{1:t}$ and $\tilde{V}^{(k)}_{1:t}$. The Discrete Fréchet distance considers all possible ordered couplings between the points of the two polygonal curves. Let an order-preserving coupling sequence $L$ be a sequence of index pairs $(i, j)$ such that $(i_1, j_1) = (1, 1)$, $(i_{|L|}, j_{|L|}) = (t, t)$, and successive pairs transition by $(+1, 0)$, $(0, +1)$, or $(+1, +1)$.

The Discrete Fréchet distance is defined as the minimum over all valid coupling sequences of the maximum Euclidean distance between coupled points:
\begin{equation}
    d_{\text{F}}\big(\tilde{P}_{1:t}, \tilde{V}^{(k)}_{1:t}\big) = \min_{L} \max_{(i,j) \in L} \big\| \tilde{p}_i - \tilde{v}^{(k)}_j \big\|_2
\end{equation}
In practice, this distance is computed efficiently utilizing dynamic programming. 

\paragraph{3. Prefix failure and Normalization:}
At each progression step $t$, the geometric failure is evaluated against the closest expert prefix:
\begin{equation}
    d^*(t) = \min_{\tilde{V} \in \tilde{\mathcal{E}}} d_{\text{F}}\big(\tilde{P}_{1:t}, \tilde{V}_{1:t}\big)
\end{equation}
To account for the natural geometric variance of the specific object's class, $d^*(t)$ is standardized. Let $\mu$ and $\sigma$ be the empirical mean and standard deviation of the pairwise Fréchet distances evaluated among the full-length resampled experts in $\tilde{\mathcal{E}}$. The normalized failure score at step $t$ is formulated as:
\begin{equation}
    \hat{d}(t) = \frac{d^*(t) - \mu}{\sigma + \epsilon}
\end{equation}
where $\epsilon$ is a small constant ensuring numerical stability. The final per-step scores form a trajectory of geometric divergence tracking the severity of the failure over the execution sequence.

\subsubsection{Soft Dynamic Time Warping (Soft-DTW) Formulation}

The Soft-DTW baseline detects anomalies by assessing the global spatio-temporal dissimilarity between a full observed trajectory and a set of nominal expert trajectories. By utilizing dynamic time warping, this approach is highly resilient to temporal scaling and non-linear shifts in execution speed.

\paragraph{1. Trajectory Preprocessing:}
Let the observed trajectory be $P = \{p_1, \dots, p_N\}$ and the set of expert trajectories be $\mathcal{E} = \{V^{(1)}, \dots, V^{(K)}\}$. To ensure computational tractability and enable batched parallel execution on GPU hardware, all trajectories are linearly downsampled to a maximum fixed length $T_{max}$ (if their original length exceeds it) and zero-padded.

\paragraph{2. Soft-DTW Distance:}
For a given test trajectory $P$ and an expert trajectory $V$ (both of length $T \leq T_{max}$), we define the pairwise Euclidean distance matrix $\Delta \in \mathbb{R}^{T \times T}$, where $\Delta_{i,j} = \|p_i - v_j\|_2^2$. Let $\mathcal{A}(T, T)$ represent the set of all valid binary alignment matrices (paths) between two sequences of length $T$. The standard DTW distance minimizes the inner product $\langle A, \Delta \rangle$ over all $A \in \mathcal{A}$.

Soft-DTW introduces a smoothing parameter $\gamma > 0$ and replaces the non-differentiable $\min$ operator with a soft-minimum, defined for a set of values $\{x_i\}$ as:
\begin{equation}
    \min{}^\gamma(x_1, \dots, x_n) = -\gamma \log \sum_{i=1}^{n} e^{-x_i/\gamma}
\end{equation}
The Soft-DTW distance between $P$ and $V$ aggregates the costs of all possible alignment paths:
\begin{equation}
    d_{\text{sDTW}}(P, V) = \min{}^\gamma_{A \in \mathcal{A}(T, T)} \langle A, \Delta \rangle
\end{equation}
This softening ensures that the distance metric is less brittle to localized noise compared to the standard, strictly optimal DTW path.

\paragraph{3. Minimum Expert Distance and Normalization:}
The raw trajectory failure is quantified as the minimum Soft-DTW distance to any expert demonstration in the nominal set:
\begin{equation}
    d^*(P) = \min_{V \in \mathcal{E}} d_{\text{sDTW}}(P, V)
\end{equation}
To determine statistical significance, this raw distance is standard-normalized. Let $\mu$ and $\sigma$ be the empirical mean and standard deviation of all pairwise Soft-DTW distances computed exclusively within the expert set $\mathcal{E}$:
\begin{equation}
    \hat{d}(P) = \frac{|d^*(P) - \mu|}{\sigma + \epsilon}
\end{equation}
where $\epsilon$ is a small scalar added for numerical stability. The resulting scalar $\hat{d}(P)$ serves as the final failure score for the evaluated trajectory.

\subsection{Evaluation Metrics}
\label{sec:eval_metrics}

We assess our framework in two stages: evaluating failure scores independently of decision thresholds, and evaluating the operational end-to-end binary failure detection. Unlike some prior works that evaluate failure detection solely at the episode level, our predictions are evaluated at a granular step level, where labels $y \in \{0, 1\}$ correspond to nominal behavior and task failure, respectively. Let $TP$, $TN$, $FP$, and $FN$ denote the number of true positives, true negatives, false positives, and false negatives.

\paragraph{Raw Failure Scoring.}
To evaluate the intrinsic discriminative power of the continuous anomaly scores $S_t$, we report the Area Under the Precision-Recall Curve (AUPR) and the maximum Matthews Correlation Coefficient (MCC). We favor AUPR because it is particularly informative under strong class imbalance and is sensitive to the precision-recall trade-off. Furthermore, to capture the optimal threshold-independent balance between precision and recall, we report the maximum MCC achieved across all possible thresholds. MCC evaluates the quality of binary classifications on skewed distributions, yielding a value between $-1$ and $1$ (with $0$ indicating random correlation and $1$ indicating perfect correlation):
\begin{equation}
    \text{MCC} = \frac{TP \times TN - FP \times FN}{\sqrt{(TP+FP)(TP+FN)(TN+FP)(TN+FN)}}
\end{equation}

\paragraph{End-to-End Detection Pipeline.}
To evaluate the practical deployment readiness of the complete pipeline, we assess the binary decisions produced by applying the Conformal Prediction (CP) thresholds. This measures the entire system's ability to safely flag critical failures while remaining robust to benign variations. We report the True Positive Rate (TPR), True Negative Rate (TNR), and the threshold-calibrated MCC:
\begin{equation}
    \text{TPR} = \frac{TP}{TP+FN}, \quad \text{TNR} = \frac{TN}{TN+FP}
\end{equation}
These metrics isolate distinct operational requirements: TPR (Recall) evaluates system safety by measuring the proportion of failures successfully intercepted, while TNR (Specificity) measures operational efficiency by quantifying the ability to ignore benign noise without triggering false alarms.

\section{Additional results}

\subsection{Threshold-independent results per task}
\label{app:raw_per_task}
Table~\ref{tab:raw_per_task} details Figure~\ref{fig:raw_scoring} per task for all nine methods, and adds AUROC.
 
\paragraph{Our two scores.} TPC is the best method on every task in AUPR and maximum MCC, except AUPR on Dish storing, where logpZO is higher (0.670 vs.\ 0.639). It outperforms GnnT on all six tasks in both metrics, including the table-setting task of the Watchdog dataset (AUPR 0.963 vs.\ 0.800) for which 100 expert demonstrations are available, so the number of demonstrations alone does not explain the gap. Waste sorting is the only task where GnnT comes close: its maximum MCC is within 0.02 of TPC (0.531 vs.\ 0.551), its AUROC is higher (0.854 vs.\ 0.762, on all 10 seeds) and it reaches the higher recall after calibration (Table~\ref{tab:e2e_per_task}), while TPC keeps a much higher AUPR (0.658 vs.\ 0.497, on all 10 seeds). We attribute this to the continuous motion of the belt: initial states never repeat, which penalizes a direct comparison to stored demonstrations more than a model that predicts object motion from the observed past.
 
\paragraph{Comparison with baselines.} GnnT outperforms every baseline on dataset averages but not on every task: logpZO is better on Dish storing (AUPR 0.670 vs.\ 0.548), AP-RMS on Vegetable sorting (AUPR 0.711 vs.\ 0.690) and AP-Fr\'echet on Groceries sorting in maximum MCC (0.527 vs.\ 0.497). No baseline, however, is strong across tasks. Trajectory-matching baselines are the most task-dependent: AP-RMS reaches an AUPR of 0.711 on Vegetable sorting, but all three fall below the chance level on Waste sorting (0.139--0.166 vs.\ a failure-step prevalence of 0.22), and they are the least stable across seeds (standard deviation up to 0.373 AUPR for AP-Fr\'echet on the Watchdog table-setting task), since their score depends on which demonstrations form the reference set. Because they compare each object independently, they cannot represent relational errors such as swapped placements. Density (logpZO, lopO), reconstruction (AE-Recon) and visual (FIDeL) baselines are stable across seeds (standard deviation at most 0.054 AUPR) but remain task-dependent: logpZO is the strongest baseline on Dish storing (0.670) yet drops to 0.189 on Groceries sorting, and FIDeL goes from 0.475 to 0.210 on the same two tasks.
 
\begin{table}[t]
\centering
\caption{Threshold-independent results per task: AUPR, maximum MCC and AUROC (mean\,$\pm$\,std over 10 seeds). Dataset averages are computed per seed over the tasks of the dataset, then averaged over seeds (as in Fig.~\ref{fig:raw_scoring}). The chance level of AUPR is the fraction of failure steps. Best in bold, second underlined. $^{\dagger}$FIDeL: 9 seeds on Table-setting (Watchdog).}
\label{tab:raw_per_task}
\resizebox{\linewidth}{!}{%
\begin{tabular}{l cccc c cc c cc}
\toprule
 & \multicolumn{4}{c}{BotFails} & & \multicolumn{2}{c}{Watchdog} & & \multicolumn{2}{c}{Dataset average} \\
\cmidrule{2-5}\cmidrule{7-8}\cmidrule{10-11}
Method & Table-setting & Dish storing & Veg. sorting & Groc. sorting & & Table-setting & Waste sorting & & BotFails & Watchdog \\
\midrule
\multicolumn{11}{l}{\textit{AUPR $\uparrow$}} \\
\textcolor{gray}{Chance (prevalence)} & 0.314 & 0.296 & 0.318 & 0.190 & & 0.218 & 0.224 & & 0.280 & 0.221 \\
AP-RMS & \ms{0.272}{0.013} & \ms{0.234}{0.011} & \ms{\underline{0.711}}{0.065} & \ms{0.549}{0.106} & & \ms{0.468}{0.287} & \ms{0.149}{0.001} & & \ms{0.442}{0.029} & \ms{0.309}{0.143} \\
AP-Fr\'echet & \ms{0.282}{0.032} & \ms{0.226}{0.011} & \ms{0.591}{0.155} & \ms{0.608}{0.262} & & \ms{0.527}{0.373} & \ms{0.166}{0.001} & & \ms{0.427}{0.066} & \ms{0.346}{0.187} \\
Soft-DTW & \ms{0.247}{0.012} & \ms{0.202}{0.004} & \ms{0.309}{0.007} & \ms{0.166}{0.041} & & \ms{0.142}{0.011} & \ms{0.139}{0.000} & & \ms{0.231}{0.010} & \ms{0.140}{0.005} \\
FIDeL & \ms{0.381}{0.000} & \ms{0.475}{0.000} & \ms{0.361}{0.000} & \ms{0.210}{0.000} & & \ms{0.297}{0.000} & \ms{0.270}{0.000} & & \ms{0.357}{0.000} & \ms{0.283}{0.004} \\
logpZO & \ms{0.474}{0.010} & \ms{\mathbf{0.670}}{0.012} & \ms{0.406}{0.054} & \ms{0.189}{0.004} & & \ms{0.266}{0.004} & \ms{0.392}{0.006} & & \ms{0.435}{0.013} & \ms{0.329}{0.004} \\
lopO & \ms{0.327}{0.026} & \ms{0.405}{0.054} & \ms{0.335}{0.017} & \ms{0.175}{0.011} & & \ms{0.188}{0.021} & \ms{0.225}{0.026} & & \ms{0.310}{0.017} & \ms{0.207}{0.020} \\
AE-Recon & \ms{0.528}{0.028} & \ms{0.584}{0.037} & \ms{0.384}{0.017} & \ms{0.176}{0.014} & & \ms{0.341}{0.014} & \ms{0.319}{0.023} & & \ms{0.418}{0.012} & \ms{0.330}{0.015} \\
TPC (ours) & \ms{\mathbf{0.775}}{0.014} & \ms{\underline{0.639}}{0.023} & \ms{\mathbf{0.939}}{0.005} & \ms{\mathbf{0.890}}{0.063} & & \ms{\mathbf{0.963}}{0.024} & \ms{\mathbf{0.658}}{0.016} & & \ms{\mathbf{0.811}}{0.016} & \ms{\mathbf{0.810}}{0.013} \\
GnnT (ours) & \ms{\underline{0.631}}{0.040} & \ms{0.548}{0.053} & \ms{0.690}{0.100} & \ms{\underline{0.623}}{0.090} & & \ms{\underline{0.800}}{0.077} & \ms{\underline{0.497}}{0.042} & & \ms{\underline{0.623}}{0.049} & \ms{\underline{0.648}}{0.051} \\
\midrule
\multicolumn{11}{l}{\textit{Maximum MCC $\uparrow$}} \\
AP-RMS & \ms{0.202}{0.035} & \ms{0.062}{0.038} & \ms{\underline{0.541}}{0.130} & \ms{0.476}{0.091} & & \ms{0.407}{0.247} & \ms{0.118}{0.021} & & \ms{0.320}{0.027} & \ms{0.262}{0.122} \\
AP-Fr\'echet & \ms{0.186}{0.014} & \ms{0.140}{0.033} & \ms{0.341}{0.110} & \ms{\underline{0.527}}{0.260} & & \ms{0.497}{0.273} & \ms{0.192}{0.012} & & \ms{0.298}{0.085} & \ms{0.344}{0.133} \\
Soft-DTW & \ms{0.196}{0.070} & \ms{0.000}{0.000} & \ms{0.193}{0.046} & \ms{0.124}{0.087} & & \ms{0.030}{0.071} & \ms{0.000}{0.000} & & \ms{0.128}{0.029} & \ms{0.015}{0.036} \\
FIDeL & \ms{0.281}{0.000} & \ms{0.215}{0.000} & \ms{0.135}{0.000} & \ms{0.032}{0.000} & & \ms{0.175^{\dagger}}{0.000} & \ms{0.124}{0.000} & & \ms{0.166}{0.000} & \ms{0.147^{\dagger}}{0.008} \\
logpZO & \ms{0.386}{0.015} & \ms{\underline{0.532}}{0.011} & \ms{0.101}{0.043} & \ms{0.028}{0.024} & & \ms{0.186}{0.009} & \ms{0.259}{0.005} & & \ms{0.262}{0.016} & \ms{0.223}{0.006} \\
lopO & \ms{0.109}{0.076} & \ms{0.128}{0.106} & \ms{0.010}{0.008} & \ms{0.017}{0.015} & & \ms{0.027}{0.036} & \ms{0.049}{0.047} & & \ms{0.066}{0.037} & \ms{0.038}{0.023} \\
AE-Recon & \ms{0.421}{0.015} & \ms{0.434}{0.038} & \ms{0.045}{0.034} & \ms{0.017}{0.021} & & \ms{0.268}{0.014} & \ms{0.176}{0.022} & & \ms{0.229}{0.013} & \ms{0.222}{0.013} \\
TPC (ours) & \ms{\mathbf{0.651}}{0.010} & \ms{\mathbf{0.614}}{0.027} & \ms{\mathbf{0.889}}{0.001} & \ms{\mathbf{0.852}}{0.067} & & \ms{\mathbf{0.888}}{0.060} & \ms{\mathbf{0.551}}{0.028} & & \ms{\mathbf{0.751}}{0.014} & \ms{\mathbf{0.719}}{0.034} \\
GnnT (ours) & \ms{\underline{0.528}}{0.051} & \ms{0.412}{0.050} & \ms{0.517}{0.115} & \ms{0.497}{0.098} & & \ms{\underline{0.731}}{0.069} & \ms{\underline{0.531}}{0.020} & & \ms{\underline{0.488}}{0.047} & \ms{\underline{0.631}}{0.040} \\
\midrule
\multicolumn{11}{l}{\textit{AUROC $\uparrow$}} \\
AP-RMS & \ms{0.470}{0.034} & \ms{0.395}{0.031} & \ms{0.828}{0.057} & \ms{0.809}{0.063} & & \ms{0.673}{0.187} & \ms{0.282}{0.006} & & \ms{0.625}{0.022} & \ms{0.477}{0.092} \\
AP-Fr\'echet & \ms{0.447}{0.039} & \ms{0.379}{0.041} & \ms{0.690}{0.091} & \ms{0.797}{0.205} & & \ms{0.677}{0.293} & \ms{0.372}{0.007} & & \ms{0.578}{0.053} & \ms{0.524}{0.148} \\
Soft-DTW & \ms{0.393}{0.040} & \ms{0.277}{0.017} & \ms{0.537}{0.014} & \ms{0.444}{0.132} & & \ms{0.234}{0.057} & \ms{0.215}{0.003} & & \ms{0.413}{0.034} & \ms{0.224}{0.028} \\
FIDeL & \ms{0.636}{0.000} & \ms{0.666}{0.000} & \ms{0.570}{0.000} & \ms{0.501}{0.000} & & \ms{0.640^{\dagger}}{0.000} & \ms{0.576}{0.000} & & \ms{0.593}{0.000} & \ms{0.605^{\dagger}}{0.010} \\
logpZO & \ms{0.732}{0.004} & \ms{\mathbf{0.846}}{0.005} & \ms{0.585}{0.070} & \ms{0.513}{0.010} & & \ms{0.633}{0.006} & \ms{0.710}{0.004} & & \ms{0.669}{0.017} & \ms{0.671}{0.004} \\
lopO & \ms{0.551}{0.046} & \ms{0.611}{0.046} & \ms{0.498}{0.025} & \ms{0.460}{0.026} & & \ms{0.441}{0.062} & \ms{0.494}{0.051} & & \ms{0.530}{0.020} & \ms{0.467}{0.045} \\
AE-Recon & \ms{0.778}{0.013} & \ms{\underline{0.799}}{0.019} & \ms{0.571}{0.015} & \ms{0.478}{0.037} & & \ms{0.683}{0.010} & \ms{0.643}{0.020} & & \ms{0.656}{0.010} & \ms{0.663}{0.012} \\
TPC (ours) & \ms{\mathbf{0.872}}{0.019} & \ms{0.777}{0.017} & \ms{\mathbf{0.974}}{0.002} & \ms{\mathbf{0.942}}{0.035} & & \ms{\mathbf{0.989}}{0.008} & \ms{\underline{0.762}}{0.016} & & \ms{\mathbf{0.891}}{0.011} & \ms{\underline{0.876}}{0.008} \\
GnnT (ours) & \ms{\underline{0.822}}{0.019} & \ms{0.777}{0.034} & \ms{\underline{0.836}}{0.060} & \ms{\underline{0.815}}{0.054} & & \ms{\underline{0.938}}{0.017} & \ms{\mathbf{0.854}}{0.013} & & \ms{\underline{0.812}}{0.023} & \ms{\mathbf{0.896}}{0.012} \\
\bottomrule
\end{tabular}}
\end{table}
 
\subsection{End-to-end results per task}
\label{app:e2e_per_task}
Table~\ref{tab:e2e_per_task} details Figure~\ref{fig:E2E_results} per task, with TPR and TNR. After calibration, TPC is the best method on five of the six tasks, with an MCC between 0.519 (Dish storing) and 0.819 (Vegetable sorting). On Waste sorting, AP-Fr\'echet (0.553) is ahead of both TPC (0.470) and GnnT (0.449), which are tied (TPC ahead on 5 of 10 seeds); GnnT trades a lower TNR (0.876 vs.\ 0.897) for a higher TPR (0.587 vs.\ 0.551). Trajectory matching benefits the most from calibration: its calibrated MCC is far above its maximum MCC (e.g., AP-Fr\'echet on Waste sorting, 0.553 vs.\ 0.192), because the time-varying band absorbs the drift of prefix-based scores along the episode, which a single global threshold cannot do. GnnT benefits less, and its calibrated MCC is below that of AP-Fr\'echet on both dataset averages (0.347 vs.\ 0.452 and 0.571 vs.\ 0.630).
 
\begin{table}[t]
\centering
\caption{End-to-end detection after conformal calibration (modulation $s_1$, averaged over $\alpha\in[0.80,0.95]$): calibrated MCC, TPR and TNR per task (mean\,$\pm$\,std over 10 seeds). Best MCC in bold, second underlined. TPR and TNR are not highlighted: a high value of one alone is not informative (a detector that flags everything has TPR\,=\,1).}
\label{tab:e2e_per_task}
\resizebox{\linewidth}{!}{%
\begin{tabular}{l cccc c cc c cc}
\toprule
 & \multicolumn{4}{c}{BotFails} & & \multicolumn{2}{c}{Watchdog} & & \multicolumn{2}{c}{Dataset average} \\
\cmidrule{2-5}\cmidrule{7-8}\cmidrule{10-11}
Method & Table-setting & Dish storing & Veg. sorting & Groc. sorting & & Table-setting & Waste sorting & & BotFails & Watchdog \\
\midrule
\multicolumn{11}{l}{\textit{Calibrated MCC $\uparrow$}} \\
AP-RMS & \ms{0.375}{0.126} & \ms{0.134}{0.058} & \ms{0.636}{0.088} & \ms{0.271}{0.107} & & \ms{0.706}{0.043} & \ms{0.295}{0.034} & & \ms{0.354}{0.050} & \ms{0.501}{0.027} \\
AP-Fr\'echet & \ms{\underline{0.446}}{0.040} & \ms{0.359}{0.140} & \ms{\underline{0.656}}{0.192} & \ms{0.348}{0.189} & & \ms{\underline{0.708}}{0.169} & \ms{\mathbf{0.553}}{0.022} & & \ms{\underline{0.452}}{0.078} & \ms{\underline{0.630}}{0.090} \\
Soft-DTW & \ms{0.285}{0.071} & \ms{0.037}{0.059} & \ms{0.560}{0.152} & \ms{0.085}{0.189} & & \ms{0.067}{0.145} & \ms{0.249}{0.049} & & \ms{0.242}{0.053} & \ms{0.158}{0.069} \\
FIDeL & \ms{0.072}{0.076} & \ms{0.363}{0.025} & \ms{0.204}{0.036} & \ms{0.241}{0.010} & & \ms{0.253^{\dagger}}{0.018} & \ms{0.113}{0.016} & & \ms{0.220}{0.014} & \ms{0.177^{\dagger}}{0.019} \\
logpZO & \ms{0.061}{0.063} & \ms{0.330}{0.035} & \ms{0.179}{0.063} & \ms{0.127}{0.050} & & \ms{0.411}{0.007} & \ms{0.246}{0.024} & & \ms{0.174}{0.019} & \ms{0.329}{0.013} \\
lopO & \ms{0.066}{0.082} & \ms{0.021}{0.042} & \ms{0.017}{0.035} & \ms{0.000}{0.000} & & \ms{-0.010}{0.023} & \ms{-0.011}{0.016} & & \ms{0.026}{0.017} & \ms{-0.010}{0.017} \\
AE-Recon & \ms{0.111}{0.107} & \ms{\underline{0.501}}{0.022} & \ms{0.125}{0.057} & \ms{0.015}{0.045} & & \ms{0.384}{0.015} & \ms{0.086}{0.035} & & \ms{0.188}{0.029} & \ms{0.235}{0.018} \\
TPC (ours) & \ms{\mathbf{0.595}}{0.031} & \ms{\mathbf{0.519}}{0.094} & \ms{\mathbf{0.819}}{0.042} & \ms{\mathbf{0.765}}{0.114} & & \ms{\mathbf{0.813}}{0.086} & \ms{\underline{0.470}}{0.047} & & \ms{\mathbf{0.675}}{0.029} & \ms{\mathbf{0.641}}{0.040} \\
GnnT (ours) & \ms{0.316}{0.159} & \ms{0.213}{0.105} & \ms{0.422}{0.149} & \ms{\underline{0.436}}{0.086} & & \ms{0.693}{0.063} & \ms{0.449}{0.070} & & \ms{0.347}{0.068} & \ms{0.571}{0.057} \\
\midrule
\multicolumn{11}{l}{\textit{TPR $\uparrow$}} \\
AP-RMS & \ms{0.556}{0.165} & \ms{0.382}{0.123} & \ms{0.721}{0.176} & \ms{0.388}{0.283} & & \ms{0.648}{0.104} & \ms{0.396}{0.048} & & \ms{0.512}{0.100} & \ms{0.522}{0.048} \\
AP-Fr\'echet & \ms{0.412}{0.050} & \ms{0.600}{0.193} & \ms{0.810}{0.090} & \ms{0.329}{0.236} & & \ms{0.680}{0.227} & \ms{0.722}{0.037} & & \ms{0.538}{0.075} & \ms{0.701}{0.117} \\
Soft-DTW & \ms{0.369}{0.093} & \ms{0.211}{0.060} & \ms{0.729}{0.118} & \ms{0.071}{0.125} & & \ms{0.099}{0.121} & \ms{0.300}{0.033} & & \ms{0.345}{0.060} & \ms{0.200}{0.050} \\
FIDeL & \ms{1.000}{0.000} & \ms{0.531}{0.065} & \ms{0.329}{0.101} & \ms{0.305}{0.057} & & \ms{0.313^{\dagger}}{0.063} & \ms{0.366}{0.054} & & \ms{0.541}{0.029} & \ms{0.340^{\dagger}}{0.048} \\
logpZO & \ms{1.000}{0.001} & \ms{0.952}{0.013} & \ms{0.810}{0.218} & \ms{0.803}{0.261} & & \ms{0.699}{0.092} & \ms{0.788}{0.038} & & \ms{0.891}{0.076} & \ms{0.743}{0.054} \\
lopO & \ms{0.506}{0.260} & \ms{0.014}{0.034} & \ms{0.002}{0.005} & \ms{0.000}{0.000} & & \ms{0.002}{0.005} & \ms{0.021}{0.029} & & \ms{0.130}{0.061} & \ms{0.011}{0.015} \\
AE-Recon & \ms{0.996}{0.006} & \ms{0.656}{0.090} & \ms{0.415}{0.110} & \ms{0.118}{0.045} & & \ms{0.526}{0.060} & \ms{0.203}{0.054} & & \ms{0.546}{0.039} & \ms{0.364}{0.032} \\
TPC (ours) & \ms{0.701}{0.111} & \ms{0.728}{0.055} & \ms{1.000}{0.000} & \ms{0.766}{0.169} & & \ms{0.952}{0.048} & \ms{0.551}{0.050} & & \ms{0.799}{0.048} & \ms{0.752}{0.036} \\
GnnT (ours) & \ms{0.433}{0.236} & \ms{0.158}{0.121} & \ms{0.438}{0.203} & \ms{0.269}{0.088} & & \ms{0.796}{0.101} & \ms{0.587}{0.093} & & \ms{0.325}{0.084} & \ms{0.691}{0.069} \\
\midrule
\multicolumn{11}{l}{\textit{TNR $\uparrow$}} \\
AP-RMS & \ms{0.817}{0.031} & \ms{0.750}{0.079} & \ms{0.888}{0.112} & \ms{0.855}{0.156} & & \ms{0.976}{0.019} & \ms{0.877}{0.031} & & \ms{0.828}{0.042} & \ms{0.926}{0.017} \\
AP-Fr\'echet & \ms{0.946}{0.014} & \ms{0.773}{0.053} & \ms{0.821}{0.274} & \ms{0.944}{0.066} & & \ms{0.972}{0.014} & \ms{0.863}{0.027} & & \ms{0.871}{0.073} & \ms{0.918}{0.016} \\
Soft-DTW & \ms{0.878}{0.030} & \ms{0.820}{0.050} & \ms{0.801}{0.243} & \ms{0.983}{0.030} & & \ms{0.956}{0.048} & \ms{0.906}{0.020} & & \ms{0.870}{0.064} & \ms{0.931}{0.022} \\
FIDeL & \ms{0.033}{0.046} & \ms{0.824}{0.046} & \ms{0.845}{0.066} & \ms{0.903}{0.035} & & \ms{0.902^{\dagger}}{0.030} & \ms{0.750}{0.054} & & \ms{0.651}{0.023} & \ms{0.821^{\dagger}}{0.037} \\
logpZO & \ms{0.023}{0.035} & \ms{0.371}{0.063} & \ms{0.353}{0.165} & \ms{0.336}{0.213} & & \ms{0.764}{0.075} & \ms{0.503}{0.069} & & \ms{0.271}{0.059} & \ms{0.633}{0.057} \\
lopO & \ms{0.561}{0.212} & \ms{0.995}{0.013} & \ms{1.000}{0.000} & \ms{1.000}{0.000} & & \ms{0.995}{0.016} & \ms{0.972}{0.038} & & \ms{0.889}{0.051} & \ms{0.983}{0.021} \\
AE-Recon & \ms{0.074}{0.101} & \ms{0.844}{0.054} & \ms{0.711}{0.065} & \ms{0.894}{0.035} & & \ms{0.860}{0.031} & \ms{0.869}{0.038} & & \ms{0.631}{0.036} & \ms{0.865}{0.025} \\
TPC (ours) & \ms{0.882}{0.048} & \ms{0.802}{0.109} & \ms{0.864}{0.037} & \ms{0.966}{0.031} & & \ms{0.916}{0.056} & \ms{0.897}{0.043} & & \ms{0.878}{0.030} & \ms{0.907}{0.029} \\
GnnT (ours) & \ms{0.868}{0.072} & \ms{0.967}{0.024} & \ms{0.923}{0.033} & \ms{0.991}{0.011} & & \ms{0.921}{0.029} & \ms{0.876}{0.011} & & \ms{0.937}{0.022} & \ms{0.898}{0.016} \\
\bottomrule
\end{tabular}}
\end{table}

\end{document}